\documentclass[conf]{new-aiaa}
\usepackage[utf8]{inputenc}

\usepackage{graphicx}
\usepackage{capt-of}
\usepackage{amsmath}
\usepackage[version=4]{mhchem}
\usepackage{siunitx}
\usepackage{longtable,tabularx}
\usepackage{mathrsfs}
\usepackage{float}
\usepackage{tabularx}
\usepackage{multirow}
\usepackage{array}

\title{A Kernel-based Resource-efficient Neural Surrogate for Multi-fidelity Prediction of Aerodynamic Field}

\author{Apurba Sarker\footnote{Graduate Research Assistant, Kevin T. Crofton Department of Aerospace and Ocean Engineering, 1600 Innovation Drive, VA 24060}, Reza T. Batley\footnote{Graduate Research Assistant, Kevin T. Crofton Department of Aerospace and Ocean Engineering, 1600 Innovation Drive, VA 24060}, Darshan Sarojini\footnote{Assistant Professor, Kevin T. Crofton Department of Aerospace and Ocean Engineering, 1600 Innovation Drive, VA 24060, AIAA Young Professional Member.}
and Sourav Saha \footnote{Assistant Professor, Kevin T. Crofton Department of Aerospace and Ocean Engineering, 1600 Innovation Drive, VA 24060, AIAA Young Professional Member.} 
}
\affil{Kevin T. Crofton Department of Aerospace and Ocean Engineering
Virginia Tech, VA, USA 24060}
\begin{document}

\maketitle

\begin{abstract}

Surrogate models provide fast alternatives to costly aerodynamic simulations and are extremely useful in design and optimization applications. This study proposes the use of a recent kernel-based neural surrogate, KHRONOS. In this work, we blend sparse high‑fidelity (HF) data with low‑fidelity (LF) information to predict aerodynamic fields under varying constraints in computational resources. Unlike traditional approaches, KHRONOS is built upon variational principles, interpolation theory, and tensor decomposition. These elements provide a mathematical basis for heavy pruning compared to dense neural networks. Using the AirfRANS dataset as a high‑fidelity benchmark and \textit{NeuralFoil} to generate low‑fidelity counterparts, this work compares the performance of KHRONOS with three contemporary model architectures: a multilayer perceptron (MLP), a graph neural network (GNN), and a physics‑informed neural network (PINN). We consider varying levels of high‑fidelity data availability (0\%, 10\%, and 30\%) and increasingly complex geometry parameterizations. These are used to predict the surface pressure coefficient distribution over the airfoil. Results indicate that, whilst all models eventually achieve comparable predictive accuracy, KHRONOS excels in resource-constrained conditions. In this domain, KHRONOS consistently requires orders of magnitude fewer trainable parameters and delivers much faster training and inference than contemporary dense neural networks at comparable accuracy. These findings highlight the potential of KHRONOS and similar architectures to balance accuracy and efficiency in multi‑fidelity aerodynamic field prediction. 
\end{abstract}

\section{Nomenclature}

{\renewcommand\arraystretch{1.0}
\noindent
\begin{longtable*}{@{}l @{\quad=\quad} l@{}}

$y$ & Ground-truth quantity \\
$\hat{y}$ & Surrogate prediction \\
$\Delta$ & Residual correction field \\

$U$ & Freestream velocity \\
$\textit{AoA}$ & Angle of attack \\
$Re$ & Reynolds number, $Re=\rho U c/\mu$ \\
$\rho$ & Fluid density \\
$\mu$ & Dynamic viscosity \\
$c$ & Airfoil chord length \\

$\bar{p}$ & Reduced (kinematic) pressure, $\bar{p}=p/\rho$ \\
$p$ & Dimensional surface pressure \\
$C_p$ & Surface pressure coefficient \\

$\boldsymbol{d}_i$ & B-spline control point coordinates \\
$\zeta$ & B-spline parameter \\
$N_{\mathrm{geom}}$ & Number of geometry control points used as inputs \\
$N_{\mathrm{Cp}}$ & Number of surface locations for $C_p$ prediction \\

$x_{\mathrm{LF}}$ & Input vector to low-fidelity surrogate \\
$x_{\Delta}$ & Input vector to residual surrogate \\

$\mathcal{K}(x)$ & KHRONOS kernel expansion mapping \\
$r$ & Rank of separable expansion \\
$k$ & Kernel grid points per input dimension \\
$g_s$ & Uniform kernel grid points on $[0,1]$ \\
$\gamma_i$ & Learnable scaling parameter for input dimension $i$ \\

$R^2$ & Coefficient of determination \\

\end{longtable*}}

\section{Introduction}\label{sec:intro}
Surrogate or reduced-order models are ``quick-to-evaluate" mathematical approximations of computationally expensive detailed models. Their efficiency is especially valuable in many-query settings, such as design space exploration~\cite{xie2022certification}, optimization~\cite{sarojini2023large, orndorff2023air,saha2025efficient,mostakim2025framework}, and uncertainty quantification~\cite{duca2018effects}, where the repetitive evaluation of computationally expensive models would otherwise be prohibitively costly. Although surrogates can target scalar~\cite{forrester2008engineering, ZHANG2021113485, koziel2013surrogate} responses or full fields~\cite{Dowell1999ROMUnsteadyAero, Benner2015projection, Halder2022NonIntrusiveROMCAE, Perron2022ManifoldAlignmentMFROM}, many engineering design-optimization workflows require \emph{full} field predictions (e.g., spatial distributions of pressure fields) to define objectives and constraints that depend on local behavior. Learning entire fields is markedly harder than predicting scalars due to the high-dimensional outputs.

The construction of surrogate models for both scalar and field quantities faces a major obstacle when the underlying response is highly nonlinear and depends on many independent parameters.
In such high-dimensional input spaces, most learning methods require substantial training data to achieve acceptable accuracy~\cite{Hou2022DimRedSurrogateModeling, Han2017WGEK, Zan2022HighDimAeroCNN}. When those data come from high-fidelity simulations, the cost can be so large that it undermines the very motivation for building a surrogate \cite{Queipo2005SurrogateBasedAnalysisOptimization}. 
Thus, using computational models that are too detailed or expensive experiments as the source data of surrogate models is impractical.

Multi-fidelity methods have been proposed as an approach to overcome this ``curse of dimensionality" that arises with high-dimensional inputs~\cite{Perron2022ManifoldAlignmentMFROM}. The key idea is to construct a surrogate by combining a few samples from an expensive high-fidelity (HF) model with many samples from a cheaper low-fidelity (LF) model, thereby achieving acceptable accuracy at reduced computational cost. However, compared to scalar quantities, the multi-fidelity combination of fields introduces additional challenges due to potential inconsistencies between the fields produced by different fidelity levels. Analyses that nominally solve the same problem: e.g., coarse vs. fine meshes, 2-D surface (vortex lattice method) vs. 3-D surface (CFD), or differing turbulence models, often yield inconsistent field representations. This inconsistency significantly complicates the process of combining high-dimensional results from multiple fidelity levels into a single surrogate model \cite{Perron2022ManifoldAlignmentMFROM}. \textbf{This work focuses on proposing an efficient neural architecture specifically to build multi-fidelity surrogates for efficient prediction of high-dimensional inconsistent aerodynamic fields in resource-constrained settings.}

Several methods have recently appeared in the literature for constructing multi-fidelity field surrogates that address heterogeneous or mismatched fidelity data using shared latent spaces, alignment, or fusion strategies. Reduced-order and manifold-alignment approaches project low- and high-fidelity solution fields into shared latent spaces and then learn cross-fidelity mappings in that space, enabling non-intrusive fusion of fields defined on different meshes or geometries \cite{Perron2022ManifoldAlignmentMFROM, Perron2020MFROM, Conti2024MultiFidelityROM, Xiao2025AdaptiveMFGappyPOD}. A large family of Gaussian-process and regression-based methods focuses on heterogeneous low-fidelity data. They include extended Co-Kriging and its variants \cite{Meng2018MultiFidelityNN}, variance-weighted sums \cite{Cheng2021VWM_MF}, local-correlation \cite{Chen2024LCWFMFS}, ensemble fusion \cite{Wang2025MDFDT}, geometric transformations \cite{HAI2024ALMFS}, and harmonic-domain or basis-adaptation strategies \cite{YOU2025102999, osti_2502152}, all designed to integrate multiple low-fidelity sources with limited high-fidelity data. More recently, deep-learning-based approaches have been proposed for high-dimensional fields, using convolutional networks\cite{ZHANG2023106354}, residual neural processes \cite{pmlr-v235-niu24d}, and multi-fidelity deep architectures \cite{Shi2024reliability, MOROZOVA2025112533, ZENG2025117657} to predict spatially distributed quantities and reliability metrics from combinations of low- and high-fidelity data. Collectively, these methods demonstrate that it is possible to build accurate multi-fidelity field surrogates even in the presence of inconsistent representations, but they also reveal a strong growth in training cost and model complexity as the dimensionality of the fields and the number of fidelity sources increase. However, these methods have a high training cost as the input dimensionality grows. Moreover, a common feature in most of these models is the need to construct an intermediate latent space to match the feature spaces of multi-fidelity domains.

To address these gaps, we propose a novel multi-fidelity approach based on \textbf{KHRONOS} (Kernel Expansion Hierarchy for Reduced-Order, Neural-Optimized Surrogates)~\cite{Batley_Saha_2025_KHRONOS}. KHRONOS represents the high-dimensional input-output mapping as tensor products of low-dimensional projected subspaces and learns only interpolation functions to represent the subspaces. This makes the surrogate fast to evaluate and largely independent of the underlying grid resolution. In addition, separability and auto-differentiability of KHRONOS enable creating multi-fidelity models without constructing an intermediate latent space, and drop-in addition into any optimization algorithm. To make comparisons fair, KHRONOS is evaluated within a common surface representation created by LF-to-HF interpolation, while remaining consistent with latent-space perspectives used in prior works \cite{saha2021hidenn,saha2024mechanistic,park2025inn_natcomms}.

This work presents a comparative study of multi-fidelity field surrogates under black-box assumptions using a \textbf{2-D airfoil aerodynamic test case}, where the low-fidelity model is a vortex-based method and the high-fidelity model is a RANS-based CFD solver; we systematically vary (a) the number of high-fidelity samples and (b) the dimensionality of the input parameterization, and assess performance in terms of prediction accuracy, training time, and inference time. 
Results show that KHRONOS matches baseline accuracy across varying HF/LF ratios with test $R^2 \ge 0.8$ and $R^2 \approx 0.90$ in Case 3, while using 94--98\% fewer parameters compared to MLP/GNN/PINN. It also reduces computational cost, achieving 2.44--3.64 ms inference and 3--15 s training per fold across cases.

The research contributions are: (1) a novel multi-fidelity surrogate modeling method for aerodynamic field quantities in a strictly non-intrusive, black-box setting, and (2) a demonstration of the practical performance of this method on 2-D airfoils, showing training time inference time reductions compared to existing state-of-the-art methods. The paper is organized as follows: Section~\ref{sec:intro} reviews background literature on multi-fidelity field surrogates; Section~\ref{sec:problem-definition} describes the 2-D airfoil test case, geometry parameterization, and datasets; Section~\ref{sec:methodology} details how the methods are applied to the airfoil problem of Section~\ref{sec:problem-definition}; Section~\ref{sec:results} presents and discusses the results; and Section~\ref{sec:conclusion} concludes with a summary and avenues for future work.

\section{Problem Definition}
\label{sec:problem-definition}

The current research uses the AirfRANS (Airfoil Reynolds-Averaged Navier–Stokes) High-Fidelity Computational Fluid Dynamics (CFD) dataset \cite{bonnet2022AirfRANS} as the high-fidelity (HF) reference, while a \textit{NeuralFoil} \cite{sharpe2025neuralfoil} surrogate model to predict and produce the corresponding low-fidelity (LF) dataset. The following sections summarize the datasets and provide a more detailed description of the problem.
\subsection{Description of the High-Fidelity Dataset (AirfRANS)}
The AirfRANS dataset provides high-fidelity, steady-state, two-dimensional, incompressible RANS solutions over airfoils in the subsonic regime. The airfoil design space is comprised of parameterized NACA (National Advisory Committee for Aeronautics) 4 and 5 digit families. These airfoils are selected, ensuring broad geometric coverage and series-specific rules intended to avoid unusable shapes. Each case combines a geometry with operating conditions drawn to reflect typical flight settings, summing up to a total of 1,000 simulations spanning Reynolds numbers ($Re$) from $2\times10^{6}\text{ to }6\times10^{6}$ and angles of attack from $-5^{\circ}\text{ to }15^{\circ}$. The simulations were performed using High-fidelity meshes generated in OpenFOAM \cite{openfoam_v2112} using a C-grid with far-field boundaries at 200 chords. The boundary layer was resolved with a first cell height of \SI{2}{\micro\metre} (target $y^{+}\approx 1$) producing \numrange{250e3}{300e3} cells per-case. Finally, the flows were solved with \texttt{simpleFoam} (\textsc{SIMPLEC}) \cite{openfoam_simplefoam} and the $k$--$\omega$ SST turbulence model \cite{menter1994twoequation}, and each run was advanced until the lift and drag coefficients converged. For learning tasks, the dataset contains full-field targets for $\bar{u}_{x}$, $\bar{u}_{y}$, $\bar{p}$ (reduced pressure), and $\nu_{t}$ (turbulent kinematic viscosity). To stabilize training and focus on the region of interest, all simulations were cropped to $[-2,4]\times[-1.5,1.5]~\si{\metre}$ and standardized using training-set statistics.

\begin{figure} [h!]
  \includegraphics[width=1.0\linewidth]{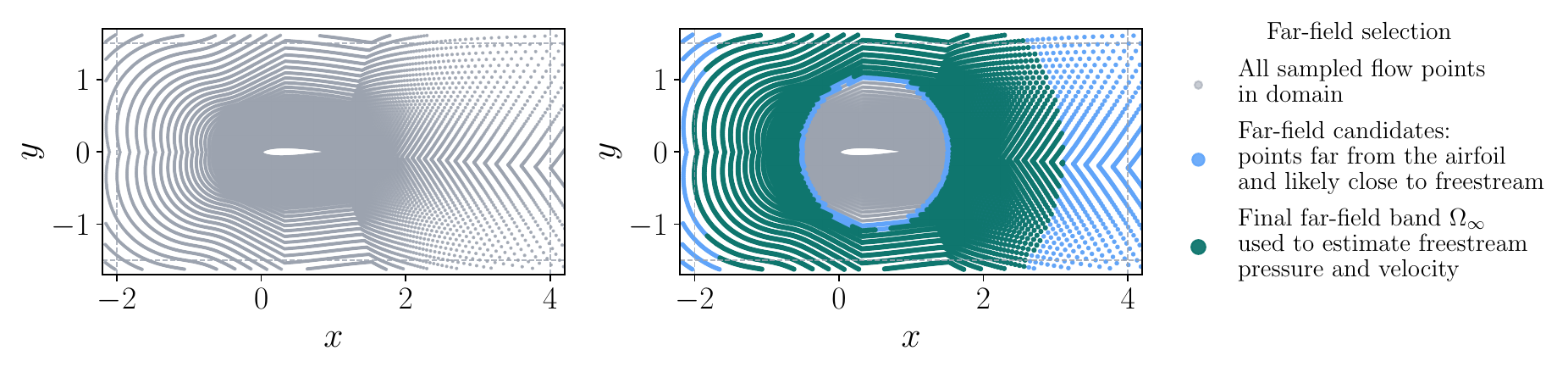}
  \caption{Selection of the far-field band $\Omega_\infty$ from internal flow samples around a representative airfoil case, used to infer the freestream reference state $(\bar{p}_\infty, U_\infty)$.}
  \label{fig:fig1}
\end{figure}

In the AirfRANS dataset, the stored pressure field corresponds to a reduced (kinematic) pressure, that is, the physical pressure divided by the density. 
\begin{equation}
  \bar{p} = \frac{p}{\rho},
\end{equation}
for which only pressure gradients appear in the incompressible RANS equations. As a result, the absolute level of reduced pressure is arbitrary. Adding a constant everywhere in the domain does not change the velocity field. Consequently, $\bar{p}$ is determined only up to an arbitrary additive constant. Any uniform shift $\bar{p} \to \bar{p} + C$ leaves the velocity field unchanged. To obtain a physically meaningful dimensional surface pressure distribution, $p(x)$, on the airfoil surface, firstly, this constant needs to be dealt with by identifying a suitable freestream reference state. This reference is estimated directly from the numerical solution by using a band of “far-field” points that are well away from the airfoil. The construction of this band is shown in Figure \ref{fig:fig1}.

For each case, the airfoil surface coordinates $(x_a, y_a)$ are first used to define the chord
length

\begin{equation}
  c = x_a^{\max} - x_a^{\min}
\end{equation}

and a geometric center

\begin{equation}
  x_c = \tfrac{1}{2}\bigl(x_a^{\max} + x_a^{\min}\bigr),
  \qquad
  y_c = \tfrac{1}{2}\bigl(y_a^{\max} + y_a^{\min}\bigr).
\end{equation}

Each internal grid point $(x_i, y_i)$ is then assigned a distance from this center,
\begin{equation}
  r_i = \sqrt{(x_i - x_c)^2 + (y_i - y_c)^2}.
\end{equation}

A far-field region $\Omega_\infty$ (blue in Figure \ref{fig:fig1}) is constructed from the set of internal points whose distance $r_i$ exceeds that of the outermost surface point by at least a fraction of the chord, so that they lie well away from the airfoil. From these candidates, the final far-field band (green) is constructed by discarding extreme outliers in radius and checking that a sufficient number of points remain. The gray points in the figure denote all internal samples. It is assumed that this far-field band represents undisturbed freestream conditions. Over this band, a simple average of the reduced pressure is taken to obtain the far-field reduced pressure, and the average of the local flow speed is used to obtain an effective freestream velocity.

Over this far-field region, the mean reduced pressure and an effective freestream speed are estimated as

\begin{equation}
  \bar{p}_\infty
  = \frac{1}{|\Omega_\infty|} \sum_{i \in \Omega_\infty} \bar{p}_i,
  \qquad
  U_\infty
  = \frac{1}{|\Omega_\infty|} \sum_{i \in \Omega_\infty}
      \sqrt{u_i^2 + v_i^2},
\end{equation}

where $(u_i, v_i)$ denote the local velocity components at the internal points. The dimensional absolute surface pressure is then reconstructed by aligning the far-field mean with a prescribed ambient pressure $P_\infty$ and rescaling by a reference density $\rho$:

\begin{equation}
  p(x) = P_\infty + \rho \bigl(\bar{p}(x) - \bar{p}_\infty\bigr),
  \label{eq:AirfRANS_pressure_reconstruction}
\end{equation}
with $\rho = 1.225~\si{kg.m^{-3}}$ and $P_\infty = 101\,325~\si{Pa}$. Finally, the surface pressure coefficient is obtained as

\begin{equation}
  C_p(x) =
  \frac{p(x) - P_\infty}{\tfrac{1}{2} \rho U_\infty^2}.
\end{equation}
In this way, both $p(x)$ and $C_p(x)$ are expressed relative to a freestream state $(P_\infty, \rho, U_\infty)$ that is inferred directly from the far-field portion of the numerical solution.

\subsection{Construction of the Low-Fidelity Dataset Using \textit{NeuralFoil}}

\textit{NeuralFoil} was trained on a very large synthetic dataset generated with \textit{XFoil} \cite{drela1989xfoil} under incompressible assumptions. \textit{NeuralFoil} covering an 18-parameter CST \cite{kulfan2008universal, kulfan2008modification} airfoil family that optionally including control-surface deflections. angles of attack ranged from \ang{-27.9} to \ang{+28.6} (median $\approx\ang{0.9}$); and chord Reynolds numbers spanned \num{0.9} to \num{2.9e12} (middle $95\%$ roughly $\numrange{1.9e3}{2.6e8}$). Freestream transition/turbulence was varied by drawing $N_{\text{crit}}\sim\mathcal{U}[0,18]$ \cite{drela2001xfoilmanual}, with forced trips applied in a minority of cases and natural transition otherwise. The model itself operated in the incompressible limit ($M_{\infty}=0$) and handled compressibility via an analytical correction at inference, enabling use up to the onset of transonic drag rise. In total, about \num{7.9e6} airfoil--condition pairs were attempted, with roughly \SI{56}{\percent} converging in \textit{XFoil} and used as supervised targets such as surface distributions and integrated coefficients.

The LF dataset was generated automatically from each CFD case by
reading the airfoil surface points $(x,y)$ and the CFD pressure column, parsing
the freestream speed $U$ and angle of attack \textit{AoA} from the filename, and ordering the airfoil coordinates by splitting into upper and lower surfaces, and sorting by $x$. The chord was computed from the extent of $x$ and used both to normalize geometry and to form a Reynolds number, $Re=\rho U c/\mu$. For the LF dataset generation, \textit{NeuralFoil} configurations used the large model in the incompressible condition setting $n_{\text{crit}}=9$ and trips at $x_{tr}=1$ on both sides. It was then evaluated at the same $(AoA,Re)$ to obtain boundary-layer edge speed ratios $u_{e}/u_{\infty}$ at 32 stations on the upper and lower surfaces. Then these were converted to surface pressure coefficient via $C_{p}=1-(u_{e}/u_{\infty})^{2}$ \cite{anderson2017fundamentals} under incompressible, inviscid outer-flow assumptions \cite{sharpe2025neuralfoil}. The $C_{p}$ values were linearly interpolated from the 32 \textit{NeuralFoil} stations to the original CFD surface points (upper vs.\ lower chosen by the sign of $y$), and converted to absolute pressure using

\begin{equation}
  p = p_{\infty} + \tfrac12 \rho U^{2}\,C_{p},
  \label{eq:pressure_from_cp}
\end{equation}

with $p_{\infty}=\SI{101325}{\pascal}$ and $\rho=\SI{1.225}{\kilogram\per\metre\cubed}$. Each LF file $(x,y,p,C_p)$ was saved as a CSV, and \textit{NeuralFoil’s} integral coefficients $(C_{L},C_{D},C_{M})$ were recorded. The Current study only focuses on $C_p$ field predictions. 

After applying the surrogate to all HF airfoils, it was observed that for 265 cases the \textit{NeuralFoil} surrogate failed to produce reasonable outputs with $R^2$ values below 0.7 and lower. Upon further investigation of these 265 cases, it was observed that these cases have a mix of complex geometry and flow conditions that are likely outside \textit{NeuralFoil's} effective training envelope. Unexpected $u_{e}/u_{\infty}$ were observed, suggesting surrogate extrapolation or low-confidence behavior in these regimes. High $C_{l}$ values were observed, consistent with cases near stall or strong separation. In such regimes, HF CFD can capture separation more faithfully, while NeuralFoil may show reduced accuracy. The remaining 735 cases were filtered and used for the training and testing of the surrogates. The filtered 265 cases were then used to evaluate the multi-fidelity gains on these challenging cases.

\begin{figure} [h!]
  \includegraphics[width=\linewidth]{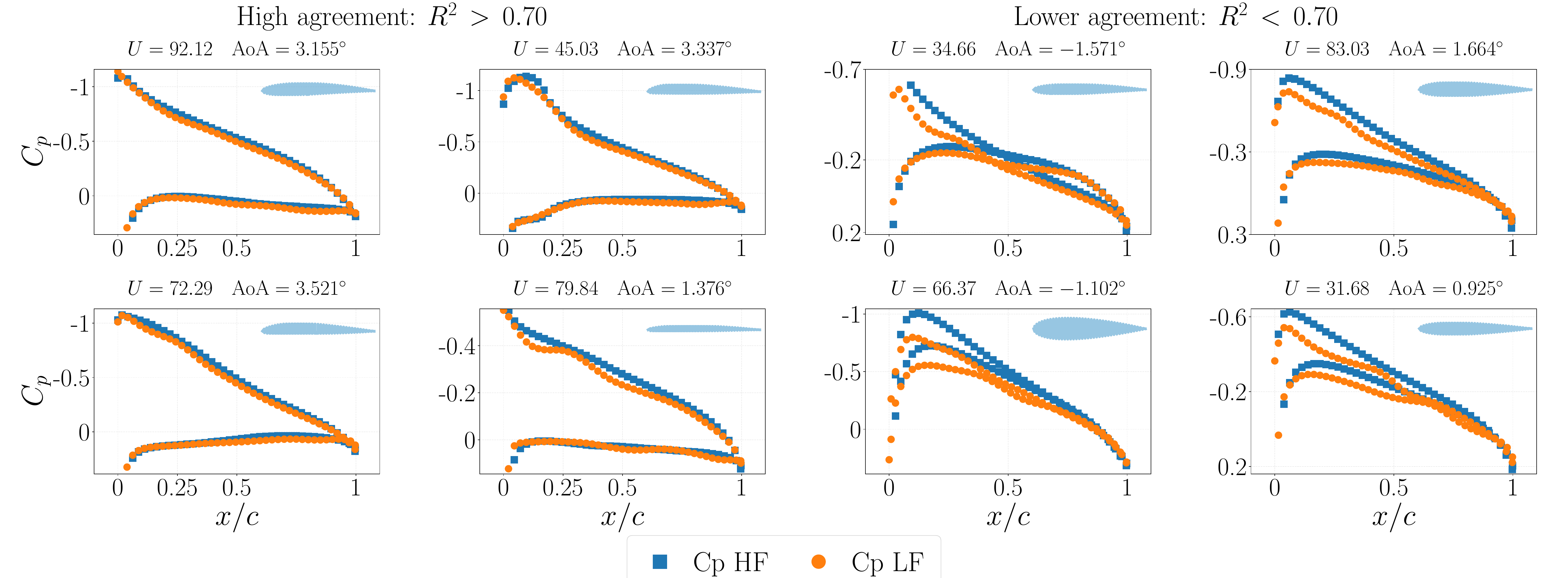}
  \caption{High fidelity (HF) vs. low fidelity (LF) $C_p$ comparisons for selected airfoils with high-agreement cases (total 735 cases) on the left and lower-agreement cases (total 265 cases) on the right.}
  \label{fig:fig2}
\end{figure}

Fig.~\ref{fig:fig2} shows the two datasets with LF $R^2$ < 0.7 in the right and LF $R^2$ > 0.7 in the left. The quality of the dataset can be analyzed from Fig.~\ref{fig:R2-bins} (see Appendix \ref{sec:neuralfoil-r2-bin}), where the number of cases in different $R^2$ ranges is listed.

\subsection{Problem Formulation and Evaluation}

The present study investigates the applicability of multi-fidelity models for predicting field quantities that vary over spatial coordinates. It proposes a \emph{novel} multi-fidelity approach based on the Kernel Expansion Hierarchy for Reduced Order, Neural Optimized Surrogates (KHRONOS) \cite{Batley_Saha_2025_KHRONOS}. Multi-layer Perceptron (MLP), Graph neural networks (GNNs), and physics-informed neural networks (PINNs) have been widely used for similar tasks. The study conducts a benchmark comparison among KHRONOS, GNNs, MLP, and PINNs to evaluate their performance across multiple aspects.

\begin{figure} [h!]
\centering
  \includegraphics[width=\linewidth]{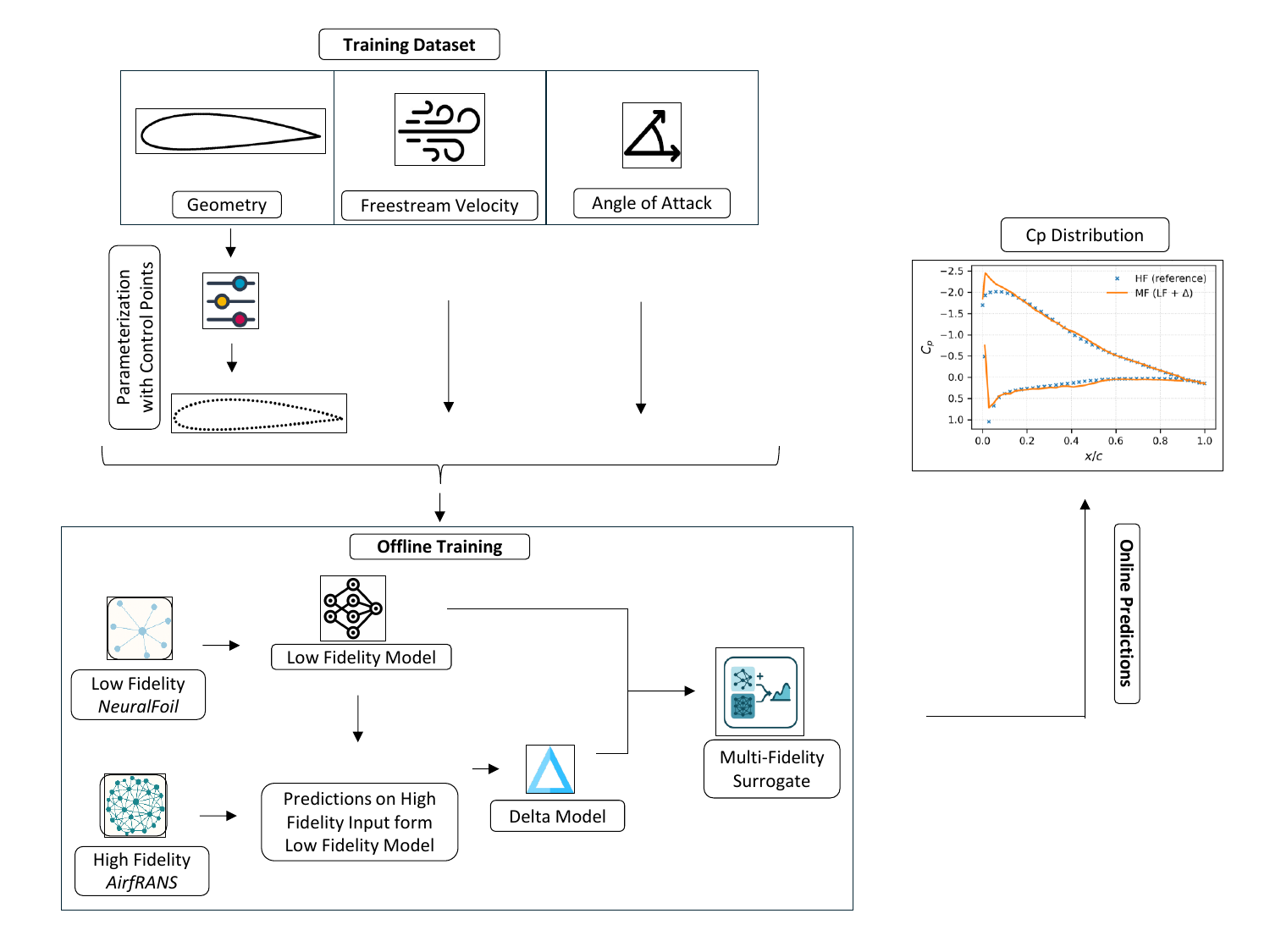}
  \caption{A schematic diagram of the multi-fidelity model development.}
  \label{fig:fig3}
\end{figure}

Fig.~\ref{fig:fig3} demonstrates the surrogate training scheme used for this study. As the training inputs, parameterized geometry (16 control points), $U$, $AoA$ are used. $U$ and $AoA$ are collected from the dataset (AirfRANS) airfoil file names. The surrogates output the $C_p$ distribution on the airfoil surface. The metrics used for comparison are inference time for predictions, training time, accuracy of the prediction using $R^2$, and number of training parameters needed for three cases (described in Table \ref{tab:case-descriptions}) with different proportions of high fidelity data. Scaling characteristics are studied by increasing the number of control points and observing the $R^2$ score and number of trainable parameters. The performance of the models is evaluated in constrained time and constrained parameters. Finally, the multi-fidelity gain of KHRONOS is demonstrated by analyzing performance with high $R^2$ LF data and low $R^2$ LF data.

\section{Computational Methodology}
\label{sec:methodology}
\subsection{Multi-fidelity Surrogate Model}

All the multi-fidelity surrogates for this study follow the same delta learning architecture \cite{meng2019composite}. The general assumption is that HF data is scarce, and LF data is abundant. So the architecture is designed to be robust enough to deal with a variable amount of HF data. As demonstrated in Fig.~\ref{fig:fig3}, offline training has two stages of training. In the first stage, an LF model is trained on the available data. Then the LF model is used to predict low fidelity predictions on the domain of HF data, $\hat{y}_{\mathrm{LF}\to\mathrm{HF}}$. Thus, a residual is calculated that is the difference between the LF model's predictions of the HF data $\hat{y}_{\mathrm{LF}\to\mathrm{HF}}$ and the ground truth from the HF dataset ${y}_\mathrm{HF}$. This difference is the delta ($\Delta$). 

\begin{equation}
  \qquad \Delta = {y}_\mathrm{HF} - \hat{y}_{\mathrm{LF}\to\mathrm{HF}}
  \label{eq:delta_definition}
\end{equation}

Then a delta model is trained to predict this delta from the LF data ${y}_\mathrm{LF}$. Thus the final multi-fidelity prediction is the summation of the $\hat{y}_\mathrm{LF}$ predictions from the LF model and $\Delta$ predictions made by the $\Delta$ model.

\begin{equation}
  \qquad \hat{y}_\mathrm{MF} = \hat{y}_\mathrm{LF} + \Delta
  \label{eq:mf_prediction}
\end{equation}

To observe the effects of increasing HF data, three separate cases are studied. The cases have different HF/LF ratios as described in Table \ref{tab:case-descriptions}. The Delta model is only used for cases 2 and 3. Case 1 surrogates are the LF models themselves. That means the case 1 surrogates do not have any delta models. They are purely the LF models without any HF correction. In this study, a grid search was implemented to find the best set of parameters for each model. The grid space is described in Appendix \ref{sec:Appendix_grid_search}. After that 5-Fold cross-validation is performed in all the best models and the mean statistics are reported. The 5-Fold cross-validation process is described in Appendix \ref{sec:K-fold}. The computational resources used for this study are listed in Table \ref{tab:hardware}.

\begin{table}[h!]
\centering
\caption{Computational resources used in this study.}
\label{tab:hardware}
\begin{tabular}{ll}
\hline
\textbf{Component} & \textbf{Specification} \\
\hline
System & Lenovo Legion 5 15AHP10 \\
CPU & AMD Ryzen 7 260 w/ Radeon 780M Graphics (8 cores, 16 threads, base 3.80 GHz) \\
GPU (integrated) & AMD Radeon 780M Graphics \\
GPU (discrete) & NVIDIA GeForce RTX 5060 (8 GB VRAM) \\
NVIDIA Driver & 573.24 \\
CUDA Version & 12.8 \\
Memory & 16.0 GB RAM (5600 MT/s) \\
\hline
\end{tabular}
\end{table}

\subsubsection{KHRONOS}
KHRONOS's representation $\mathcal{K}$ is a specific instantiation of a Separable Neural Architecture (SNA). In particular, this formulation takes a multivariate mapping and decomposes it to an additive combination of rank-$r$ individual products of one-dimensional \emph{atoms}. Each atom is a univariate basis expansion, here a quadratic B-spline expansion defined on a uniform knot vector on the unit interval.

For a $d$-dimensional input $x=(x_1,\dots,x_d)\in[0,1]^d$, the atom associated with dimension $i$ and rank $j$ is written,

\begin{align}
    \sum_s\alpha_{sij}\psi_{sij}(x_i),
\end{align}

where each $\alpha$ is a learnable coefficient and $\psi$ defines each B-spline basis function. The full KHRONOS expansion then takes the form,

\begin{align}
    \mathcal{K}(x)=\sum_{j=1}^r\prod_{i=1}^d\sum_s\alpha_{sij}\psi_{sij}(x_i).
\end{align}

\begin{figure}[h!]
    \centering
\includegraphics[width=0.65\linewidth]{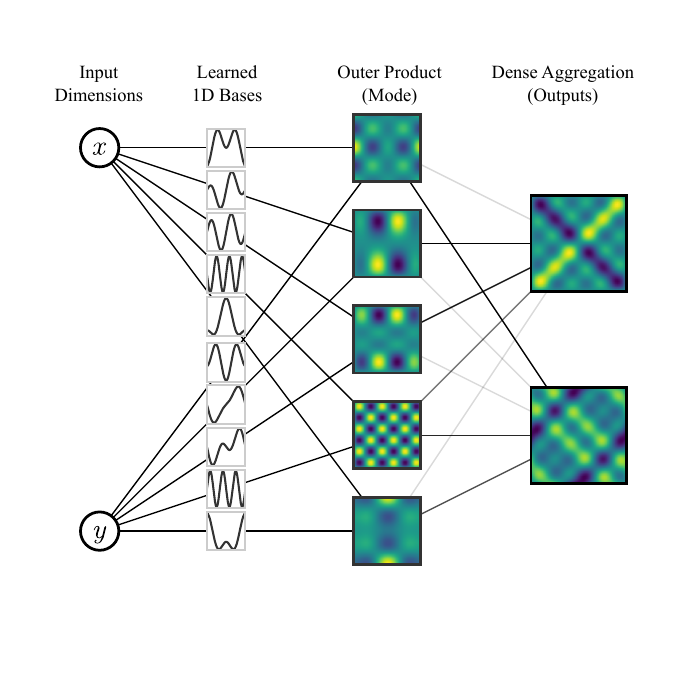}
    \caption{An illustration of KHRONOS learning a functional representation from 2-input, 2-output sampled data.}
    \label{fig:KHRONOS}
\end{figure}

The choice $\mathcal{X}=[0,1]^d$ is standard and is made without loss of generality via affine normalization. This structure is illustrated in Figure \ref{fig:KHRONOS}. A rank-5 KHRONOS instantiation learns a representation for \begin{align}
\label{eq:example}
    y=\begin{pmatrix}
    \sin(2\pi x)\sin(2\pi y)+\sin(4\pi x)\sin(4\pi y)+\sin(6\pi x)\sin(6\pi y) \\
    \cos(2\pi x)\cos(2\pi y)+\cos(4\pi x)\cos(4\pi y)+\sin(6\pi x)\sin(6\pi y)
    \end{pmatrix}
\end{align}
sampled at 3,000 points, with  30 elements per dimension. It achieves a normalized root mean square error of $0.5\%$. This example is selected to illustrate KHRONOS' ability to ascertain an efficient modal structure. It learns a single fully separable mode for each of the additive terms in \eqref{eq:example}, and that the $\sin(6\pi x)\sin(6\pi y)$ term can be shared between output heads.

This tensor product structure lends itself to models that are smooth and compact. Because each dimension is independently represented by a low-order basis, cross-dimensional interaction arises only through outer products. This inductive bias, whilst seemingly restrictive, is well-suited to those problems with low intrinsic dimensionality. It is also notable that KHRONOS is a universal approximator. Overall, this structure leads to a linear growth in parameters with increasing ambient dimensionality $d$, an extremely favorable property for training and inference, both in time and in memory, as well as in data efficiency.

In this study, a KHRONOS implementation is used in which each input component is represented by a compact quadratic spline response on a uniform grid over $[0,1]$. A set of $k$ grid points $\{g_s\}_{s=1}^{k} \subset [0,1]$ is defined, and a learnable positive scaling parameter is introduced for each input dimension so the model can adapt the width of the spline response for that variable. The scaled distance is written as
\begin{align}
  d_{is}(x_i) = |x_i - g_s| \, \gamma_i, \qquad \gamma_i > 0.
\end{align}
A quadratic B-spline activation with compact support is then evaluated at $d_{is}$. This produces a small set of localized basis responses for each input dimension. These responses are combined with learnable weights to form the per-dimension contributions for each rank component. In this way, each input dimension is modeled through a smooth localized 1D function. Interactions across dimensions are captured through the rank-$r$ separable product in $\mathcal{K}(x)$.

The KHRONOS representation is used to construct both low-fidelity and multi-fidelity predictors of the surface pressure coefficient distribution. The low-fidelity model is defined as a map from airfoil geometry and operating conditions to the pressure coefficient evaluated at a fixed set of $N_{\mathrm{Cp}}$ (81 points) surface locations. The airfoil geometry is represented by $N_{\mathrm{geom}}$ (16) sampled control points $(x,y)$ and is normalized by chord so that $x \in [0,1]$ and $y$ is expressed in chord units. The freestream speed $U$ and $AoA$ are normalized to $[0,1]$ using a min--max transformation. The flattened geometry and the two scalar flow parameters are concatenated to form the low-fidelity input.

\begin{align}
  x_{\mathrm{LF}} \in [0,1]^{d_{\mathrm{LF}}}, \qquad
  d_{\mathrm{LF}} = N_{\mathrm{geom}} + 2,
\end{align}

and a KHRONOS-based predictor is used to produce $\widehat{C}_{p}^{\mathrm{LF}} \in \mathbb{R}^{N_{\mathrm{Cp}}}$. For the LF model, rank $r=4$ is used. The LF model is trained with a peak learning rate of $3\times10^{-3}$, and $1000$ epochs.

To incorporate high-fidelity information, a residual Delta model is defined on top of the low-fidelity prediction. It uses the LF predictions to predict the HF corrections. The Delta model is trained to predict the differences between the HF values and the LF predictions. The same normalized geometry and flow variables are used, and the normalized low-fidelity $C_p$ field is appended to form the Delta input.

\begin{align}
  x_{\Delta} =
  \bigl[x_{\mathrm{LF}},\, \widehat{C}_{p}^{\mathrm{LF}}\bigr]
  \in [0,1]^{d_{\Delta}}, \qquad
  d_{\Delta} = N_{\mathrm{geom}} + 2 + N_{\mathrm{Cp}}.
\end{align}

A second KHRONOS-based predictor is then used to estimate a correction delta field $\widehat{\Delta C}_p$, and the multi-fidelity pressure estimate is obtained as

\begin{align}
  \widehat{C}_{p}^{\mathrm{MF}} =
  \widehat{C}_{p}^{\mathrm{LF}} + \widehat{\Delta C}_p .
\end{align}
For the Delta model, rank $r=6$ is used. The Delta model is trained with a peak learning rate of $10^{-3}$ and $1500$ epochs. All model hyperparameters are shown in Table \ref{tab:khronos-grid}.

In this formulation, broad trends are represented by the low-fidelity model, while systematic discrepancies are captured by the residual model using a smaller set of high-fidelity samples.

\begin{table}[H]
\centering
\caption{Selected KHRONOS hyperparameters from the grid search}
\label{tab:khronos-grid}
\begin{tabular}{lccccc}
\hline
Model & $k$ & $r$ & $\eta$ & Epochs \\
\hline
LF    & 3 & 4 & $3\times 10^{-3}$ & 1000 \\
Delta & 3 & 6 & $1\times 10^{-3}$ & 1500 \\
\hline
\end{tabular}

\vspace{0.25em}
\centering
$HF/LF$ ratio = [0,0.1,0.3],$\quad w_{\mathrm{HF}}=10$
\end{table}

\subsubsection{Multi-Layer Perceptron}
An MLP baseline is used with the same low-fidelity and residual Delta structure as the KHRONOS setup, but with standard fully connected networks instead of separable spline atoms. In the LF model, the normalized airfoil geometry is flattened and concatenated with the normalized $U$ and $AoA$. The per-case input vector is

\[
x_{\mathrm{LF}} = [\boldsymbol{x}_{\mathrm{geom}},\, U,\, \alpha]
\in \mathbb{R}^{N_{\mathrm{geom}} + 2}.
\]
with $\boldsymbol{x}_{\mathrm{geom}}$ the normalized control points, $U$ the freestream speed, $AoA$ the angle of attack.
A feedforward MLP is used to predict the normalized low-fidelity pressure distribution at 81 surface locations. In the implementation, four hidden layers are used, each with width of 256 and ReLU activations, followed by a final linear layer that outputs $N_{\mathrm{Cp}}$ (81) values. All hyperparameters are listed in Table \ref{tab:mlp-grid}

The Delta model is defined in the same spirit. The same geometry and flow inputs are used, and the normalized LF prediction is appended to form the Delta input. This augmented vector is passed through another MLP with the same depth and width, followed by a linear output layer that predicts a correction field at the $N_{\mathrm{Cp}}$ points. The final multi-fidelity estimate is formed by adding this correction to the LF prediction.

Training is carried out with Adam \cite{kingma2014adam} and a learning-rate schedule is used in both LF and Delta training.

\begin{table}[H]
\centering
\caption{Selected MLP hyperparameters from the grid search}
\label{tab:mlp-grid}
\begin{tabular}{lcc}
\hline
Hyperparameter & LF MLP & Delta MLP \\
\hline
Hidden layers & 4 & 4 \\
Hidden width & 256 & 128 \\
Activation & ReLU & ReLU \\
Optimizer & Adam & Adam \\
Peak learning rate $\eta$ & $3\times 10^{-3}$ & $1\times 10^{-3}$ \\
Epochs & 3500 & 4000 \\
\hline
\end{tabular}

\vspace{0.25em}
\centering
$HF/LF$ ratio = [0,\, 0.1,\, 0.3], $\quad w_{\mathrm{HF}}=10$
\end{table}


\subsubsection{Graph Neural Network}

A GNN baseline is used with the same low-fidelity and residual Delta multi-fidelity structure as the KHRONOS and MLP setups, but with message passing over the airfoil control-point geometry. The geometry points are treated as nodes, and each node updates its features by aggregating information from itself plus its immediate left and right neighbors, which effectively forms a chain graph along the airfoil ordering. At each layer, the local neighbor features are concatenated, then passed through a linear transform and a ReLU nonlinearity to produce the updated node embedding. This update is repeated for multiple message-passing layers. The airfoil shape is represented as an ordered set of normalized control points, and the $U$ and $AoA$ are normalized and used as inputs. An LF GNN is trained to predict the normalized low-fidelity pressure coefficient distribution at a fixed set of $N_{\mathrm{cp}} = 81$ surface locations. The hyperparameters used are described in Table \ref{tab:gnn-hparams}

A residual Delta GNN is then defined on top of the LF prediction. The same geometry and flow inputs are used, and the normalized LF pressure prediction is included to form the Delta input. A correction field is predicted at the same $N_{\mathrm{Cp}}$ locations, and the multi-fidelity output is obtained by adding the predicted correction to the LF prediction. Residual targets are formed in physical $C_p$ space with high-fidelity residuals used during Delta training.

\begin{table}[H]
\centering
\caption{Selected GNN hyperparameters from the grid search}
\label{tab:gnn-hparams}
\begin{tabular}{lcc}
\hline
Hyperparameter & LF GNN & Delta GNN \\
\hline
Message-passing layers & 4 & 4 \\
Hidden dimension & 96 & 128 \\
Output size & $N_{\mathrm{Cp}} = 81$ & $N_{\mathrm{Cp}} = 81$ \\
Optimizer & Adam & Adam \\
Peak learning rate $\eta$ & $3\times 10^{-3}$ & $3\times 10^{-3}$ \\
Epochs per fold & 3500 & 4500 \\
\hline
\end{tabular}

\vspace{0.25em}
\centering
$HF/LF ratio = [0,\, 0.1,\, 0.3],\quad w_{\mathrm{HF}}=10$
\end{table}

\subsubsection{Physics Informed Neural Network}

To complement the purely data-driven MLP multi-fidelity baseline, a physics-informed variant is introduced in which inviscid potential-flow structure is embedded into the learning problem. In this formulation, an MLP is used to represent a scalar velocity potential, $\phi_\theta$, defined over the chord-normalized airfoil coordinate system. Parameterized control points ($\boldsymbol{x}_{\mathrm{geom}}$), $U$ and $AoA$ are provided to the network, and $\phi_\theta(\boldsymbol{x}_{\mathrm{geom}},U,\alpha)$ is predicted. The associated velocity field is obtained by automatic differentiation as $\mathbf{u}_\theta=\nabla\phi_\theta$. For incompressible, irrotational flow, the potential is required to satisfy the Laplace equation,

\begin{equation}
\nabla^2\phi = 0,
\end{equation}

and this constraint is enforced by sampling interior collocation points in a padded domain surrounding the airfoil. In addition, a far-field constraint is imposed so that $\nabla\phi_\theta$ is matched to the freestream velocity vector $\mathbf{U}_\infty = U(\cos\alpha,\sin\alpha)$. For delta models, this condition is relaxed to $\nabla\phi_\theta\approx\mathbf{0}$ to encourage localized corrections. The surface pressure is recovered from Bernoulli's relation,

\begin{equation}
p_\theta = p_\infty + \frac{1}{2}\rho\left(U^2-|\nabla\phi_\theta|^2\right),
\end{equation}

and is converted to the pressure coefficient,

\begin{equation}
C_p = \frac{p_\theta-p_\infty}{\tfrac{1}{2}\rho U^2}.
\end{equation}

Training is performed by minimizing a composite objective in which supervised data agreement is blended with physics residual penalties,

\begin{equation}
\mathcal{L}
= \mathcal{L}_{\mathrm{data}}
+ \lambda_{\mathrm{pde}}\,\mathbb{E}_{\Omega}\!\left[\left(\nabla^2\phi_\theta\right)^2\right]
+ \lambda_{\mathrm{far}}\,\mathbb{E}_{\Omega_\infty}\!\left[\left\|\nabla\phi_\theta-\mathbf{U}_\infty\right\|^2\right].
\end{equation}

where $\Omega$ and $\Omega_\infty$ denote interior and far-field samples, respectively. Interior collocation points for the PDE residual are sampled uniformly within a rectangular domain spanning $x \in [-0.2,\, 1.2]$ and $y \in [-0.2,\, 1.2]$ in chord-normalized coordinates, corresponding to a padded region extending $0.2$ chord lengths beyond the airfoil. Far-field points are sampled on an annular region at a radius between $1.6$ and $2.2$ chord lengths from the airfoil center. At each training iteration, $N_{\mathrm{coll}} = 256$ interior points and $N_{\mathrm{far}} = 64$ far-field points are randomly resampled to provide stochastic coverage of the constraint regions. By constraining the network with these physically meaningful operators while retaining the same multi-fidelity structure, regularization toward dynamically consistent pressure fields is achieved for both low-fidelity and Delta predictions, and improved robustness is promoted when high-fidelity labels are sparse. The grid searched hyperparameters are similar to the MLP model, as shown in Table \ref{tab:pinn-hparams}

\begin{table}[H]
\centering
\caption{Selected PINN hyperparameters from the grid search}
\label{tab:pinn-hparams}
\begin{tabular}{lcc}
\hline
Hyperparameter & LF PINN & Delta PINN \\
\hline
Hidden layers & 4 & 4 \\
Hidden width & 256 & 128 \\
Activation & ReLU & ReLU \\
Optimizer & Adam & Adam \\
Peak learning rate $\eta$ & $3\times 10^{-3}$ & $1\times 10^{-3}$ \\
Epochs & 3500 & 4000 \\
$N_{\mathrm{coll}}$ & 256 & 256 \\
$N_{\mathrm{far}}$ & 64 & 64 \\
$\lambda_{\mathrm{pde}}$ & $10^{-1}$ & $10^{-1}$ \\
$\lambda_{\mathrm{far}}$ & $10^{-2}$ & $10^{-2}$ \\
\hline
\end{tabular}
\vspace{0.25em}

\centering
$HF\ Ratio=[0,0.1,0.3],\quad w_{\mathrm{HF}}=10$
\end{table}

\subsection{Geometry Parameterization}
There are a variety of geometry parametrizations presented in the literature for airfoils, including PARSEC, Class-Shape Transformation (CST), and Bézier, to name a few. It has been shown that the NACA 4-series, CST, and PARSEC methods are exactly equivalent to Bézier parametrization, although the reverse is not true. Since B-splines are a generalization of Bézier curves, all of the aforementioned parametrizations can be considered subsets of the B-spline parametrization \cite{rajnarayan2018universal}.\newline
In many common parametrizations, perturbing a single parameter affects the shape along the entire airfoil chord. B-splines, however, do not suffer from this limitation and allow the designer to introduce highly localized and intuitive geometry modifications. Additionally, thanks to the knot insertion capability, B-splines are well-suited for progressive design approaches, where the optimizer can incrementally refine the geometry parametrization in specific regions \cite{han2014adaptive}.\newline
A planar B-spline parametrization is used in this paper, initially to fit the existing airfoil datasets. Once the variation bounds of the control points are determined, each control point can be perturbed within its envelope to introduce geometry modifications during the shape optimization process. B-splines provide a mapping from a one-dimensional parametric space $\{\zeta \in \mathbb{R}\}$ to the two-dimensional physical space $\{\boldsymbol{X} \in \mathbb{R}^2\}$, and are expressed in the form:

\begin{equation} \label{eq:equation4}
\boldsymbol{X}(\zeta) = \sum_{i=1}^n \boldsymbol{d}_i \mathcal{N}_{i,k}(\zeta)
\end{equation}

where $\boldsymbol{d}_i$ denotes the coordinates of the control points, $n$ is the number of control points, and $\{\mathcal{N}_{i,k} : i=1,\dots,n\}$ are the normalized polynomial basis functions of order $k$. A cubic B-spline with 16 control points is used to fit the airfoils. The positions of the control points are determined by solving the following least-squares problem:

\begin{equation} \label{eq:equation5}
\min_{\{\boldsymbol{d}_i\}} \sum_{j=1}^{N} \left\| \boldsymbol{X}_j - \boldsymbol{X}(\zeta_j) \right\|_2^2
\end{equation}

where $N$ is the number of points used to represent the airfoil geometry, and $\boldsymbol{X}_j$ denotes the coordinates of the $j^{\text{th}}$ point on the airfoil from the dataset.

\begin{figure}
    \centering
   \includegraphics[width=1.00\linewidth]{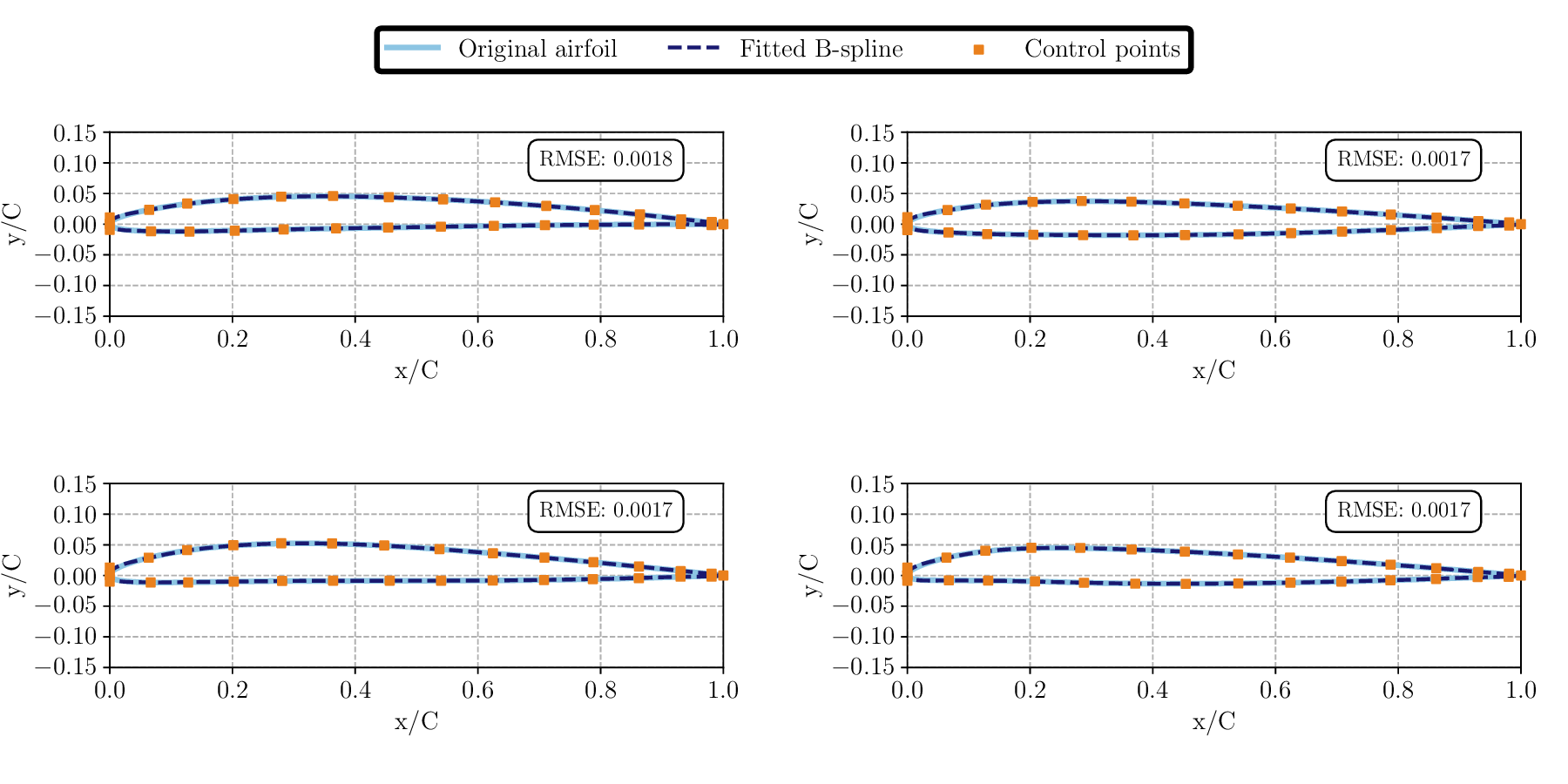}
   \caption{The four airfoils with the largest B-spline fitting errors, along with their corresponding root-mean-square error (RMSE) values. }
   \label{fig:fig5}
 \end{figure}

The B-spline is fitted to 1000 airfoils from the AirfRANS dataset. To control the AoA, two control points are fixed at the same $X$-coordinate as the leading edge control point. The B-spline is fitted to the suction side and pressure side separately, and then the two control points at the extremities of the curves are constrained to coincide, in order to maintain $\mathcal{C}^0$ continuity at the trailing edge and $\mathcal{C}^2$ continuity at the leading edge.\newline
The worst four airfoil fits are shown in Fig.~\ref{fig:fig5}, with a maximum root-mean-square error of $0.0073$. After the fitting process, the interpolated $\boldsymbol{\zeta}$ and the knot vector, which is independent of the airfoil coordinates, are used to construct the modal basis functions. Once the modal basis functions are known, the airfoil can be reconstructed using new control point coordinates according to Eq.~\ref{eq:equation5}.

\section{Numerical Results and Analysis}
\label{sec:results}
\subsection{Comparison of Computational Performance under Limited Resources}
In this section, the performance of the models is evaluated under training time, and trainable parameter constraints. For this, the grid searched Case 3 models are used along with 16 control points, $U$ and $AoA$ as input.  
\subsubsection{Computational Performance under Time Constraint}

\begin{figure}[H]
\centering
  \includegraphics[width=0.6\linewidth]{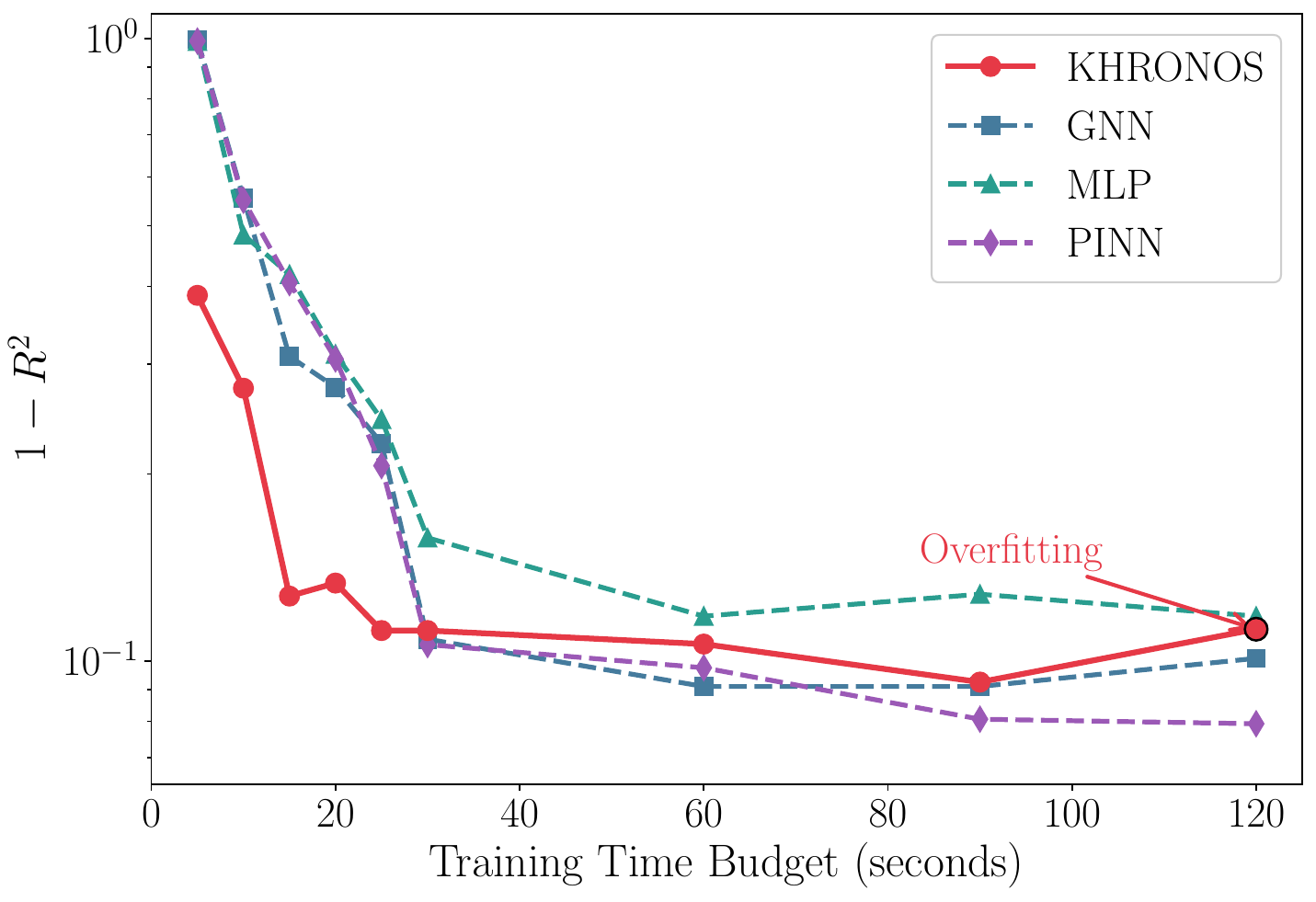}
  \caption{Prediction error $(1 - R^{2})$ as a function of the training time budget for the four surrogate models. KHRONOS is observed to reach low prediction error much faster than the GNN, MLP, and PINN baselines, reflecting its more parameter–efficient kernel representation and resulting advantage under strict time constraints.}
  \label{fig:timebudget}
\end{figure}

In Fig.~\ref{fig:timebudget}, the trade-off between training time budget and predictive accuracy is illustrated for the four surrogate models. The prediction error, measured as $1 - R^{2}$, is plotted against the available training time in seconds. To obtain this curve, each surrogate was trained from scratch for a set of predefined wall-clock budgets, using the same fixed train-test split and unchanged architectural and optimization settings for all budgets. Training was stopped once the elapsed time reached the budget, and $R^{2}$ was then evaluated on the held-out test set, with no warm-starting or checkpoint reuse between time points. For very small time budgets, all baseline models (GNN, MLP, and PINN) are characterized by high prediction errors close to unity, whereas substantially lower errors are obtained by KHRONOS. As the training time budget is increased, the error of all methods is reduced, but the most rapid decrease is consistently observed for KHRONOS. Around 15 seconds, KHRONOS already attains errors near $0.1$, while the alternative models still exhibit noticeably larger errors. For larger budgets, beyond approximately 60 seconds, the errors of all methods approach a similar low level, with KHRONOS maintaining a slight advantage. 
This advantage of KHRONOS over the GNN, MLP, and PINN baselines is attributed to the underlying kernel–based architecture and the associated training procedure. In KHRONOS, each input feature is first embedded in a low–dimensional space through a small number of compactly supported kernel functions (quadratic B–splines in the present implementation), and these one–dimensional feature maps are subsequently combined through separable tensor–product modes and superposition. As a result, the target field is represented directly in a structured function space with built–in smoothness and partition of unity properties, rather than being approximated indirectly by compositions of dense linear layers and pointwise nonlinear activations, as in standard MLP, GNN, and PINN architectures. This design yields a surrogate with far fewer trainable parameters and a lower per–epoch computational complexity, so that more effective optimization steps are completed within the same wall–clock time budget.

\subsubsection{Computational Performance under Parameter Constraint}

\begin{figure}[h!]
\centering
  \includegraphics[width=0.6\linewidth]{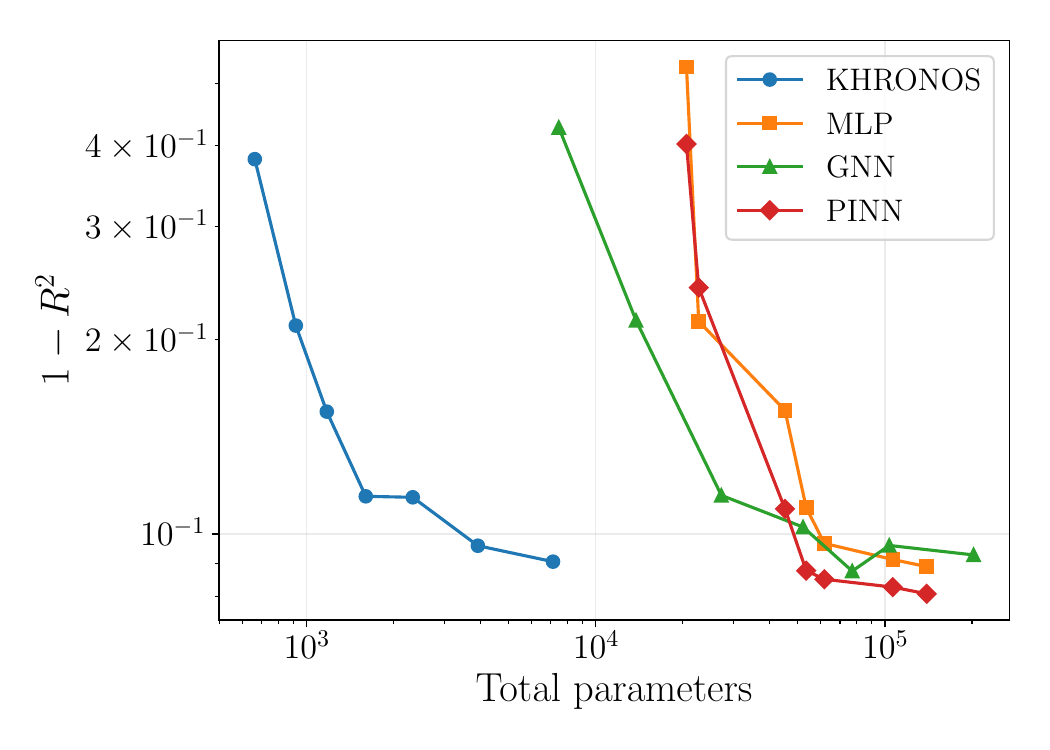}
  \caption{Prediction error ($1 - R^2$) versus total model parameters for KHRONOS, MLP, GNN, and PINN on a log--log scale. Lower values indicate better agreement with the reference data, highlighting the accuracy--complexity trade-offs across model families.}
  \label{fig:parameterbudget}
\end{figure}

Figure \ref{fig:parameterbudget} suggests a clear accuracy–complexity trade-off across all four model families that, as total parameters increase, the error $1-R^2$ generally decreases. To construct this plot, a controlled parameter sweep is performed for each model family by varying the primary architecture-size hyperparameters that govern model capacity, while the data split and training protocol are kept fixed across configurations. Each configuration is trained from scratch and is evaluated on the same held-out test set, and the total number of trainable parameters is computed directly from the instantiated model weights to provide the x-axis of the accuracy–complexity curve. On the log–log axes, the trends look close to power-law-like improvements, which is common when performance gains taper gradually with scale. KHRONOS appears to achieve strong accuracy at comparatively low parameter counts, indicating good parameter efficiency in this regime. MLP, GNN, and PINN improve with scale as well, but their curves seem to flatten at larger sizes, implying diminishing returns once the models are already near their best observed $R^2$ range. Overall, the plot indicates that if deployment constraints matter, KHRONOS may offer the most favorable performance per parameter.

\subsection{Comparison of Computational Performance with No Resource Limitation}

\subsubsection{Comparison of Overall Model Performance}

\begin{figure} [h!]
\centering
  \includegraphics[width=0.85\linewidth]{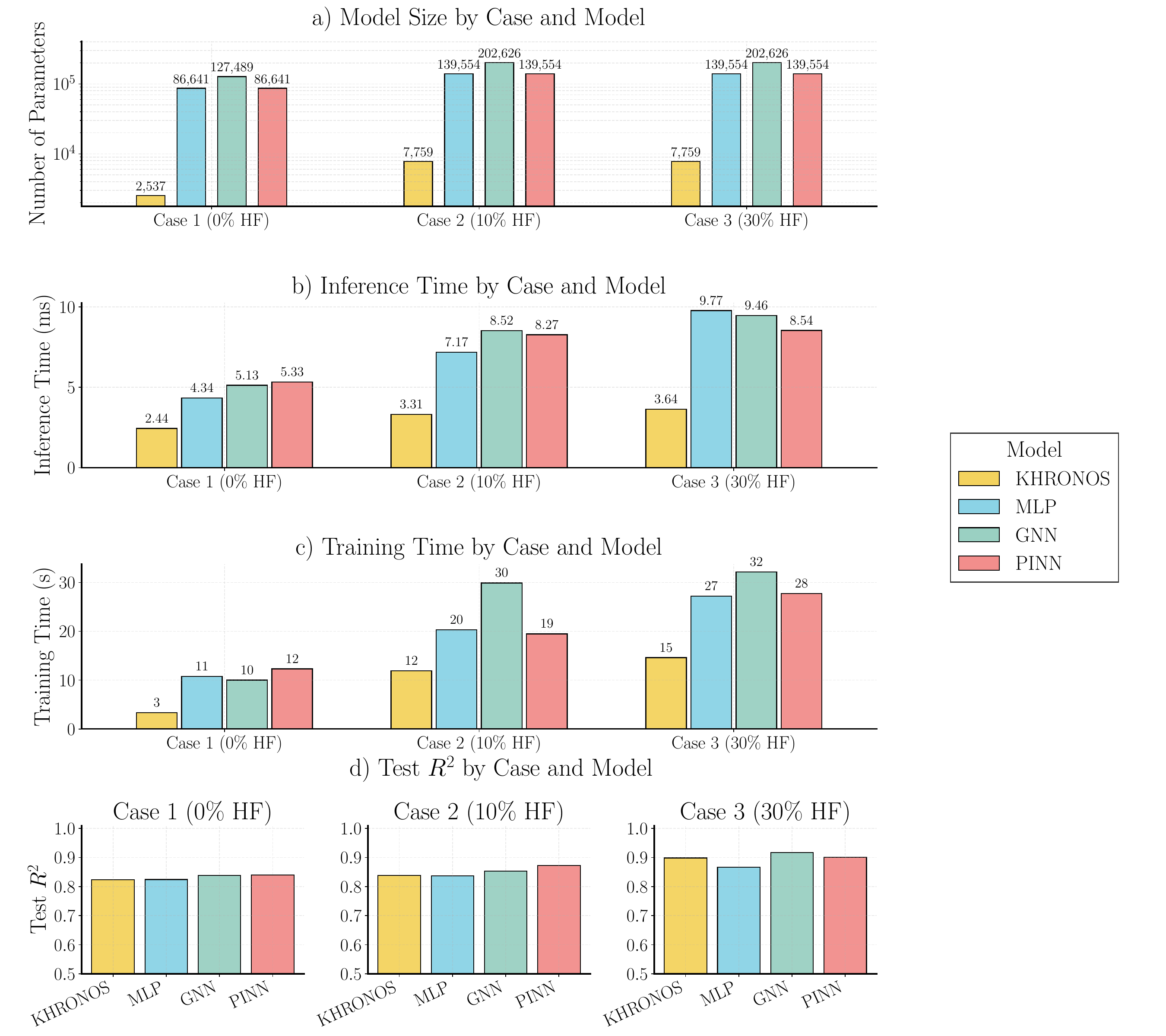}
  \caption{Comparison of the proposed KHRONOS surrogate with three neural baselines (MLP, GNN, and PINN) over the three problem configurations (Cases~1--3). Panels~(a)--(c) report, respectively, the number of trainable parameters, the inference time per evaluation, and the training time per cross-validation fold, while panel~(d) shows the corresponding test-set coefficient of determination \(R^2\).}
  \label{fig:best_models}
\end{figure}

Figure~\ref{fig:best_models} compares the proposed KHRONOS surrogate with three neural baselines (MLP, GNN, and PINN) across the three problem configurations (Cases~1--3). Panels~(a)--(c) report the model complexity and computational cost in terms of number of trainable parameters, inference time per sample, and training time per cross-validation fold, respectively, while panel~(d) shows the corresponding test performance in terms of the \(R^2\). These results are obtained after performing 5-Fold cross-validation of the models and Figure \ref{fig:best_models} reports the mean statistics. More details on the cross-validation process can be found in Appendix \ref{sec:K-fold}.

\begin{table}[H]
\centering
\caption{Dataset composition for three training cases with varying HF availability.}
\label{tab:case-descriptions}
\begin{tabular}{lccccc}
\hline
Case & Total LF data & LF training data & HF training data & Test data & HF/LF ratio \\
\hline
1 & \multirow{3}{*}{735} & \multirow{3}{*}{588} & 0   & \multirow{3}{*}{147} & 0\%  \\
2 &                      &                      & 59  &                      & 10\% \\
3 &                      &                      & 176 &                      & 30\% \\
\hline
\end{tabular}
\end{table}

In Fig.~\ref{fig:best_models}(a), KHRONOS is two to three orders of magnitude more compact than the purely data-driven networks. In Case~1, KHRONOS uses only \(2,537\) parameters, whereas the MLP and PINN each require \(86{,}641\) parameters, and the GNN uses \(127{,}489\). For the more demanding multi-fidelity settings in Cases~2 and~3, the capacity of all models is increased; KHRONOS grows moderately to \(7,759\) parameters, while the MLP and PINN each reach \(139{,}554\) parameters and the GNN \(202{,}626\). These differences follow directly from the architectural choices: KHRONOS employs a low-rank tensor-product basis, whereas the MLP and PINN rely on deep fully-connected layers with width in the hundreds, and the GNN introduces additional message-passing layers whose parameter count scales with the number of geometric control points.

The computational trends in Figs.~\ref{fig:best_models}(b) and~(c) mirror the parameter counts. KHRONOS achieves the lowest or second-lowest inference latency in all three cases (e.g., \(2.44\,\text{ms}\) in Case~1 and \(3.64\,\text{ms}\) in Case~3), and consistently exhibits the smallest training times (from \(3\,\text{s}\) in Case~1 to \(15\,\text{s}\) in Case~3). The MLP and PINN incur substantially higher inference and training costs, reflecting the cost of multiple large dense layers. The GNN is additionally penalised by the neighbor-aggregation operations required in each message-passing step, which explains why its training and inference times exceed those of KHRONOS. The PINN shares the same forward architecture as the MLP but optimises a loss that includes PDE residuals and boundary-condition terms evaluated at collocation points; the associated higher-order automatic differentiation leads to training times that are comparable to or higher than those of the MLP even though their inference costs are similar.

Despite these large differences in complexity and cost, the predictive performance in Fig.~\ref{fig:best_models}(d) is comparable across models. All methods achieve high test accuracy, with \(R^2 \gtrsim 0.8\) in all three cases. In Case~1, the four models yield similar scores around \(R^2 \approx 0.82\text{--}0.84\). As the problem becomes more challenging in Cases~2 and~3, the GNN and PINN obtain slightly higher \(R^2\) values, but KHRONOS also improves and attains \(R^2 \approx 0.90\) in Case~3, only marginally below the best-performing baselines. Overall, the figure demonstrates that KHRONOS attains accuracy comparable to the much larger data-driven and physics-informed networks while using orders of magnitude fewer parameters and substantially lower training and inference times.

\subsubsection{Scaling Characteristics for 16, 32, and 64 control points}

\begin{figure} [h!]
\centering
  \includegraphics[width=0.7\linewidth]{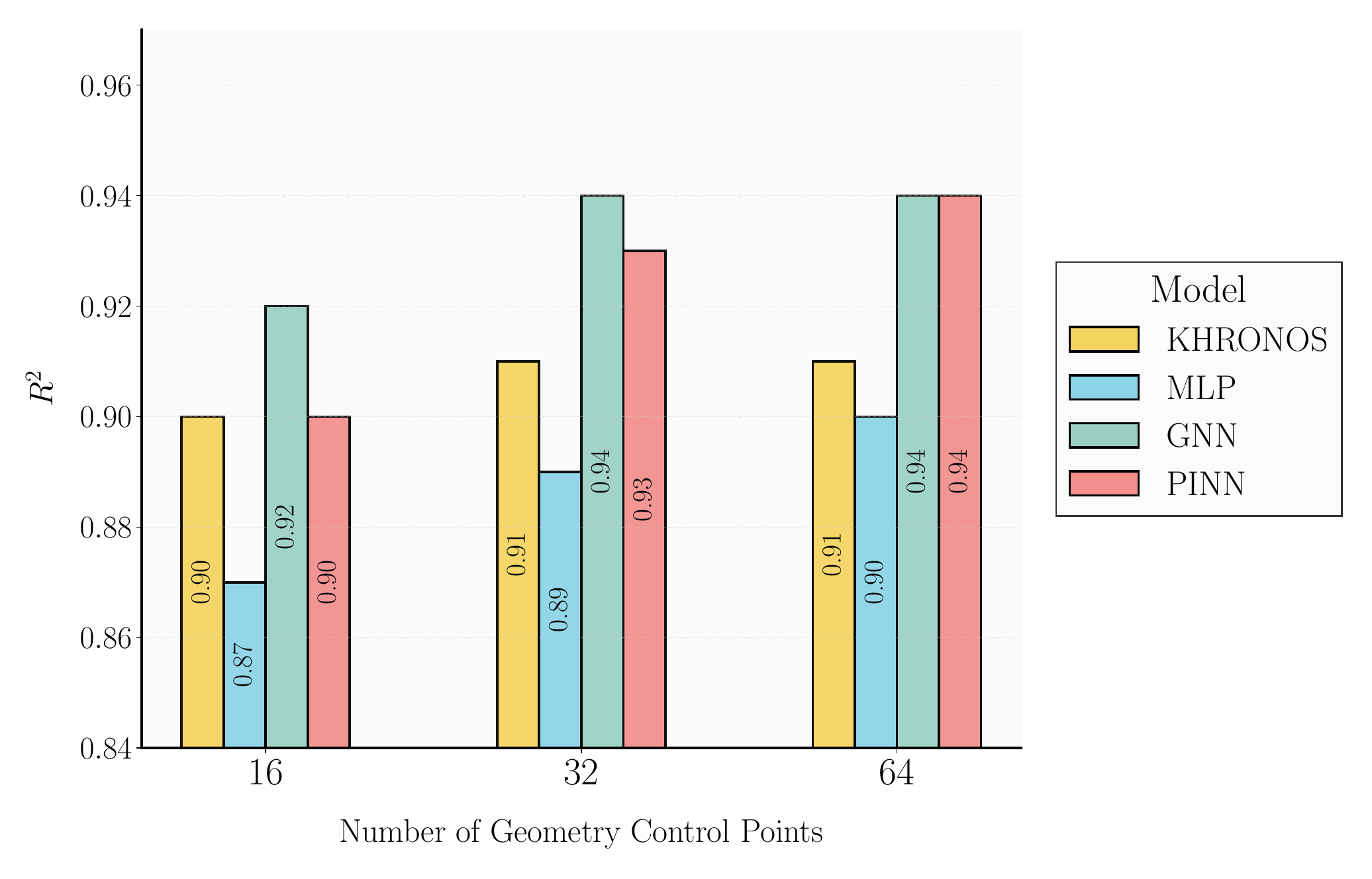}
  \caption{Scaling of predictive accuracy with the number of geometry control points for the four surrogate models. The coefficient of determination $R^{2}$, evaluated on the test set, is shown for KHRONOS, MLP, GNN, and PINN. All models achieve comparable accuracy, with $R^{2}$ values in the range $0.87$–$0.94$ as the geometric resolution is increased from $16$ to $64$ control points.}
  \label{fig:R2scaling}
\end{figure}

Figure~\ref{fig:R2scaling} investigates how the predictive accuracy changes when the geometric resolution of the airfoil parameterization is refined from 16 to 64 control points. For this experiment, the Case 3 (HF/LF ratio = 0.3) settings are used for each of the models. Across all resolutions, the four models achieve high coefficients of determination, with test \(R^2\) confined to the narrow range \(0.87\leq R^2 \leq 0.94\).  For 16 control points, KHRONOS and the PINN already reach \(R^2 = 0.90\), the MLP attains \(R^2 = 0.87\), and the GNN achieves the best score with \(R^2 = 0.92\).  Increasing the number of control points to 32 and 64 yields only modest improvements: KHRONOS saturates around \(R^2 \approx 0.91\), the MLP improves to \(R^2 = 0.89\) and \(0.90\), and the GNN and PINN reach \(R^2 = 0.94\) at 64 points.  The small gain in accuracy relative to the substantial increase in geometric degrees of freedom indicates that the models already operate in a regime where the dominant error sources are data noise, model bias in the flow physics, and limited training coverage, rather than insufficient geometric resolution.  Put differently, refining the airfoil parameterization beyond roughly 32 control points does not provide commensurate additional information to the surrogates for predicting the $C_p$ field.

\begin{figure}[h!]
\centering
  \includegraphics[width=0.6\linewidth]{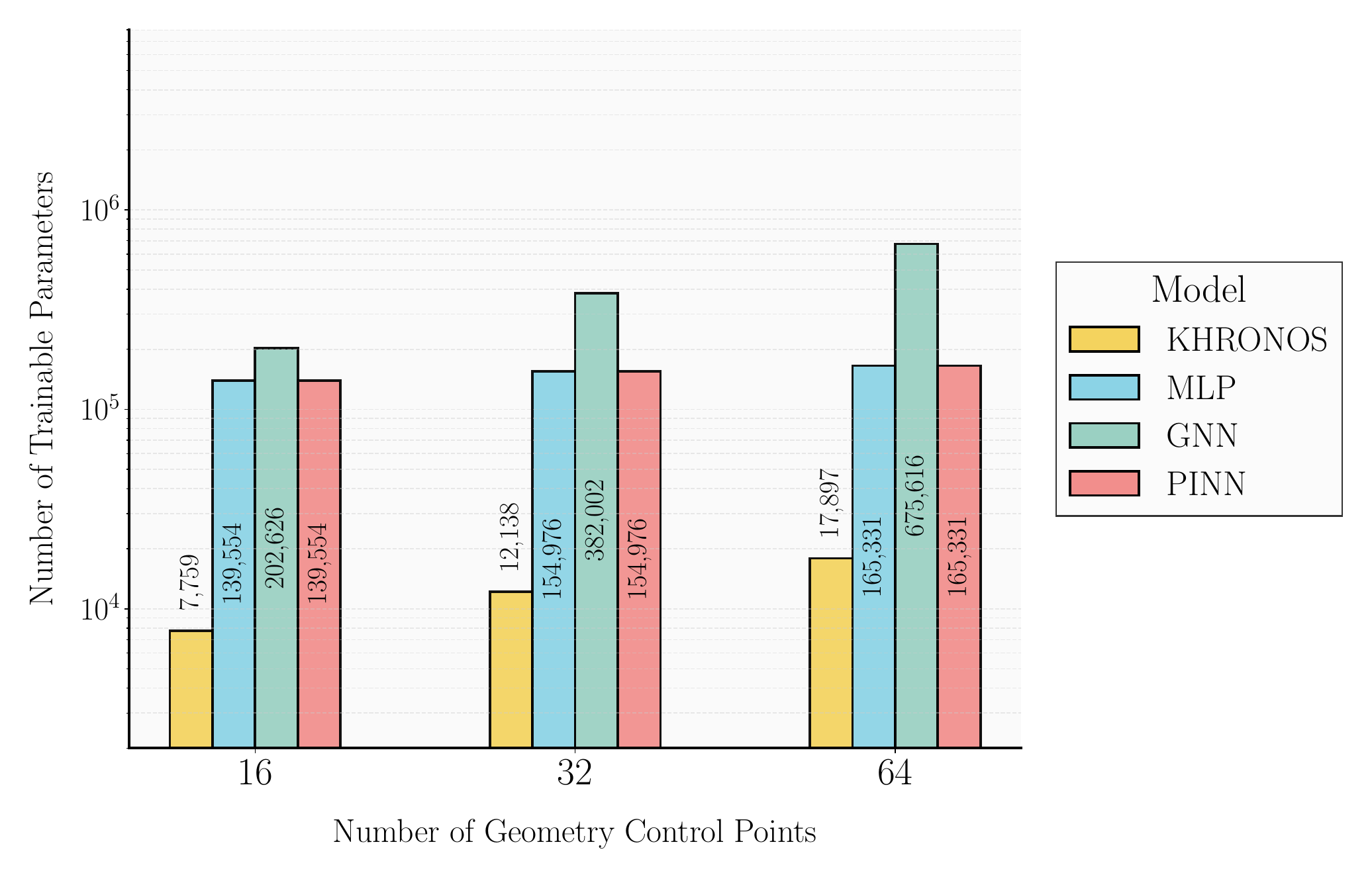}
  \caption{Scaling of the number of trainable parameters with the number of geometry control points for the four surrogate models (logarithmic vertical axis). While the parameter counts of the MLP, GNN, and PINN baselines lie in the range \(\boldsymbol{1.4\times10^{5}}\)–\(\boldsymbol{6.8\times10^{6}}\) and grow rapidly with the geometric resolution, the KHRONOS architecture requires only \(\boldsymbol{7.8\times10^{3}}\)–\(\boldsymbol{1.8\times10^{4}}\) parameters over the same range. Combined with Fig. 8, this indicates that similar levels of accuracy are achieved by KHRONOS with roughly one to two orders of magnitude fewer trainable parameters, depending on the baseline.}
  \label{fig:parameterscaling}
\end{figure}

The corresponding parameter-scaling behaviour, shown in Figure~\ref{fig:parameterscaling}, is markedly different and highlights the impact of the architectural choices. The number of trainable parameters in KHRONOS grows gently from 7,759 (16 control points) to 12,138 (32) and 17,897 (64), consistent with its low-rank tensor-product representation in which the parameter count scales essentially linearly with the number of geometry elements.  By contrast, the dense neural baselines are two to three orders of magnitude larger: the MLP and PINN already contain 139,554 parameters at 16 points and increase to 154,976 and 165,331 parameters at 32 and 64 points, respectively.  This growth stems from the widening of the input and intermediate layers as additional control-point features are concatenated, so that the cost of the first few matrix–vector products scales with the geometric discretisation.  The GNN shows the steepest scaling, with its parameter count rising from 202,626 at 16
control points to 382,002 and 675,616 at 32 and 64 points.  In the current implementation, this behaviour reflects the use of control-point-dependent message-passing blocks and edge networks, whose weight matrices increase in size with the number of nodes and connectivity in the geometry graph.

Taken together, these results demonstrate that modest increases in geometric resolution lead to only incremental improvements in \(R^2\), whereas the number of trainable parameters and hence memory footprint and computational cost  grow substantially for the conventional MLP, GNN, and PINN architectures. KHRONOS, in contrast, maintains a nearly resolution-independent parameter budget while delivering \(R^2\) values within \(0.03\) of the best-performing baseline, thereby achieving a significantly more favourable trade-off between accuracy and model complexity.

\subsection{KHRONOS's Performance on Error-prone LF Data and Complex Geometries}

\begin{figure}[h!]
\centering
  \includegraphics[width=1.0\linewidth]{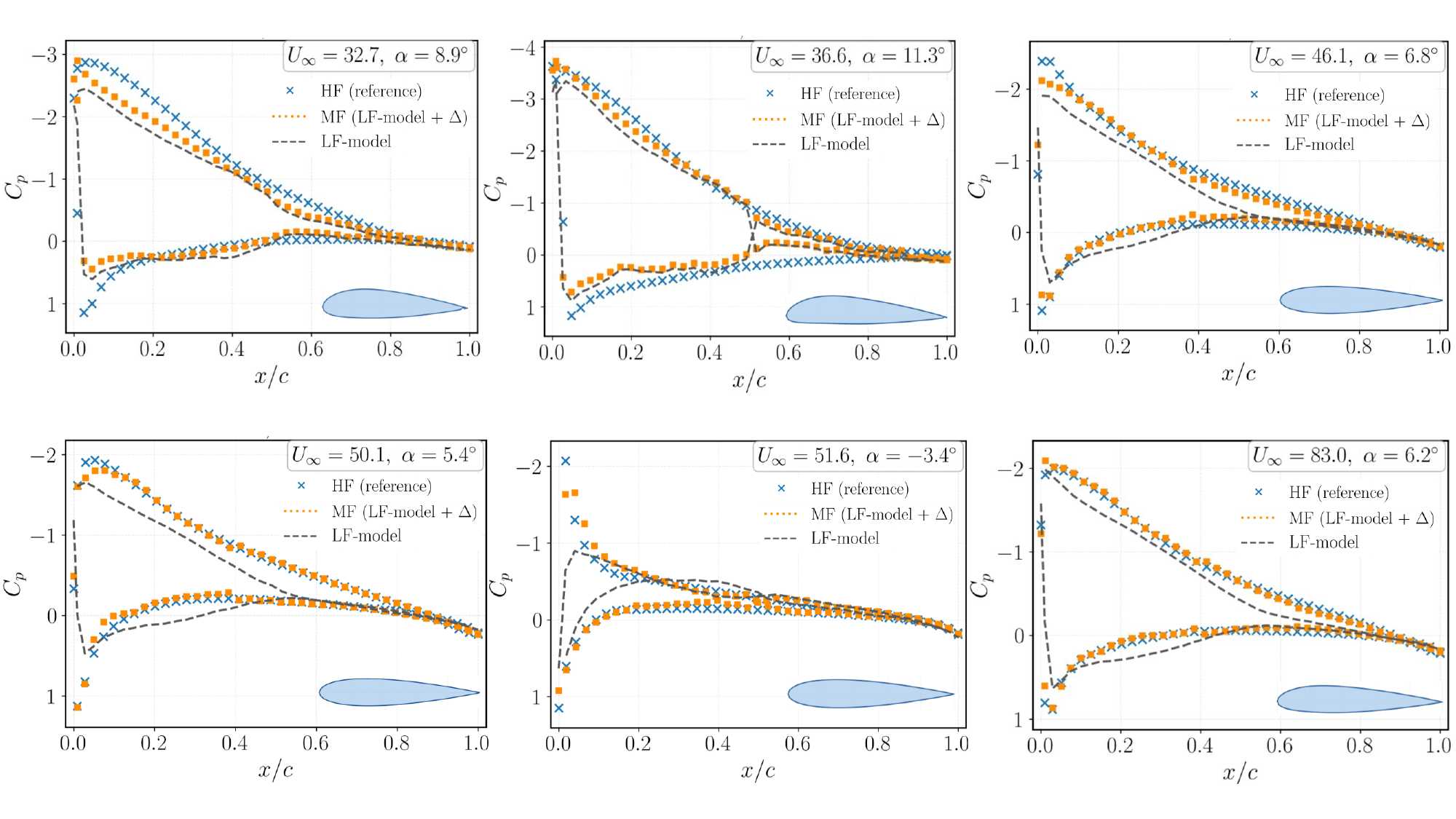}
  \caption{Representative high–accuracy test cases for the multi–fidelity (MF) KHRONOS model. Surface pressure coefficient distributions $C_p$ over several airfoil geometries are shown for different free–stream velocities $U$ and angles of attack (AoA). The MF prediction (LF~$+\Delta$, orange line) is seen to closely follow the high–fidelity (HF) reference data (blue crosses) on both upper and lower surfaces, indicating that $R^{2}$ values above $0.8$ are consistently achieved across varying shapes and operating conditions.}
  \label{fig:cp_highR2}
\end{figure}

\begin{figure}[H]
  \includegraphics[width=1.0\linewidth]{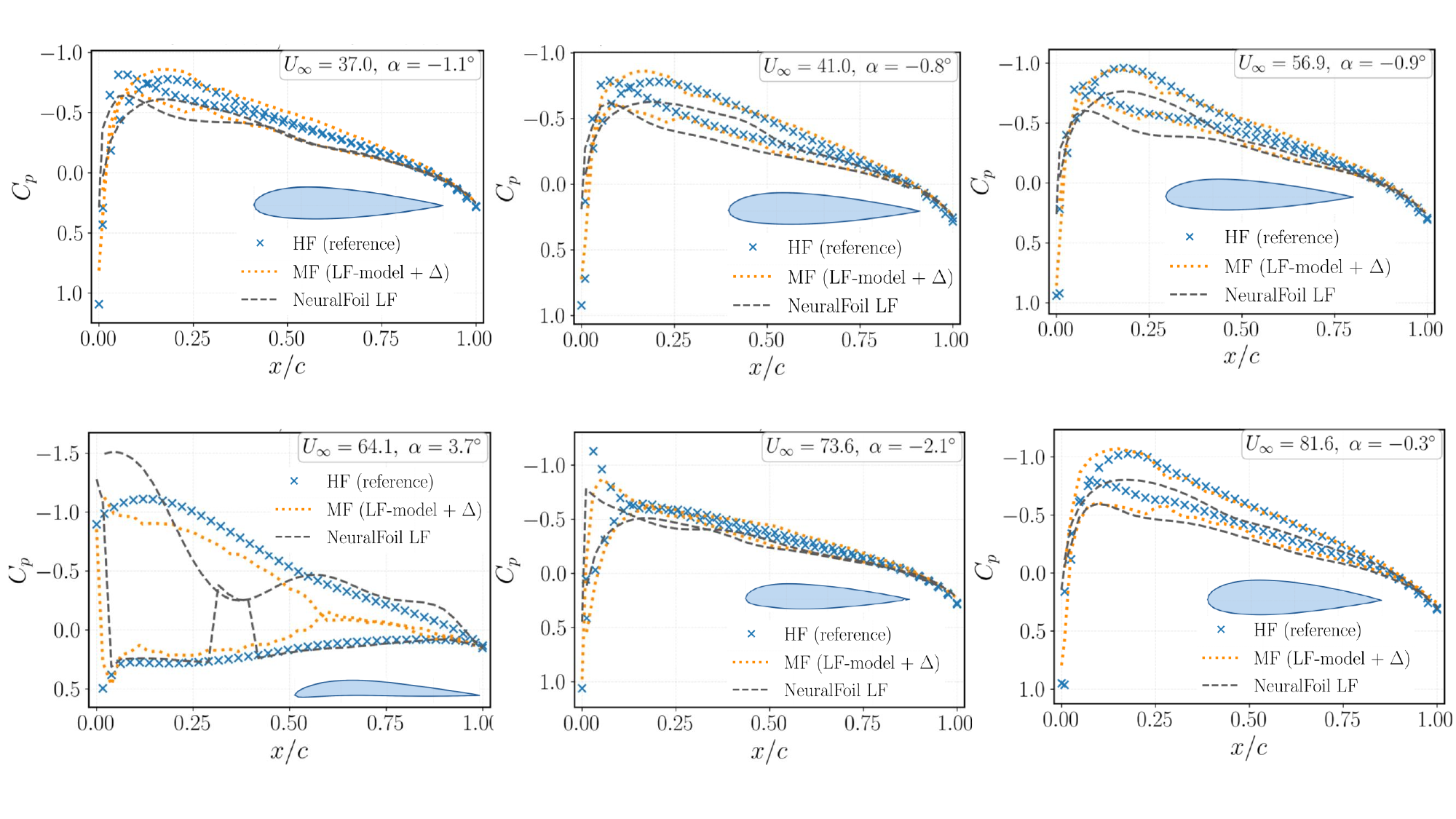}
  \caption{Representative low--$R^{2}$ LF cases comparing the standalone \textit{NeuralFoil} low--fidelity (LF) model and the corresponding multi-fidelity KHRONOS corrections. For a range of airfoil geometries, $U$, and $AoA$, \textit{NeuralFoil} (dashed gray line) can underpredict the leading-edge suction peak. The MF KHRONOS prediction (LF--model~$+\Delta$, orange line) provides partial correction of the suction peak and improves the pressure recovery region, resulting in closer agreement with the HF data for these examples.}
  \label{fig:cp_lowR2}
\end{figure}

Figures~\ref{fig:cp_highR2} and~\ref{fig:cp_lowR2} illustrate how the multi–fidelity KHRONOS model predicts surface $C_p$ distributions over a range of airfoil geometries, free–stream velocities $U$, and angles of attack (AoA). For these results the Case 3 MF KHRONOS surrogate is used with 16 control point parametrization. In the high–$R^{2}$ LF set examples from Fig.~\ref{fig:cp_highR2}, the MF KHRONOS curves (LF~$+\Delta$) align closely with the high–fidelity (HF) data for both upper and lower surfaces. The location and magnitude of the leading–edge suction peak, the subsequent suction plateau, and the pressure recovery toward the trailing edge are captured accurately for all shown combinations of $U$ and $AoA$. These cases demonstrate that, when the underlying flow is relatively well behaved, the correction learned by KHRONOS preserves good agreement with HF and yields $R^{2}$ values around $0.9$ across diverse airfoil shapes.

In contrast, Fig.~\ref{fig:cp_lowR2} focuses on cases for which the stand–alone \textit{NeuralFoil} low–fidelity model exhibits low predictive quality ($R^{2}<0.7$). A clear trend is observed that the low–$R^{2}$ set is dominated by flows with pronounced leading–edge suction heads and steep suction plateaus. In these regimes, the \textit{NeuralFoil} LF prediction (gray dashed line) tends to underpredict the depth of the suction peak and often misplaces it in the streamwise direction, which leads to a substantial mismatch with the HF reference, especially over the front $20$–$40\%$ of the chord. The MF KHRONOS prediction (LF–model~$+\Delta$) shows multi-fidelity gain while reconstructing the HF pressure distribution very closely.

The learned correction restores both the amplitude and the shape of the suction head and significantly improves the pressure recovery region, so that the orange curves lie much closer to the blue HF data than the dashed LF model in all panels. Even in the challenging examples, where \textit{NeuralFoil} does not capture the suction peak, KHRONOS produces a more consistent pressure distribution and reduced pointwise error.

\begin{figure}[H]
\centering
  \includegraphics[width=0.6\linewidth]{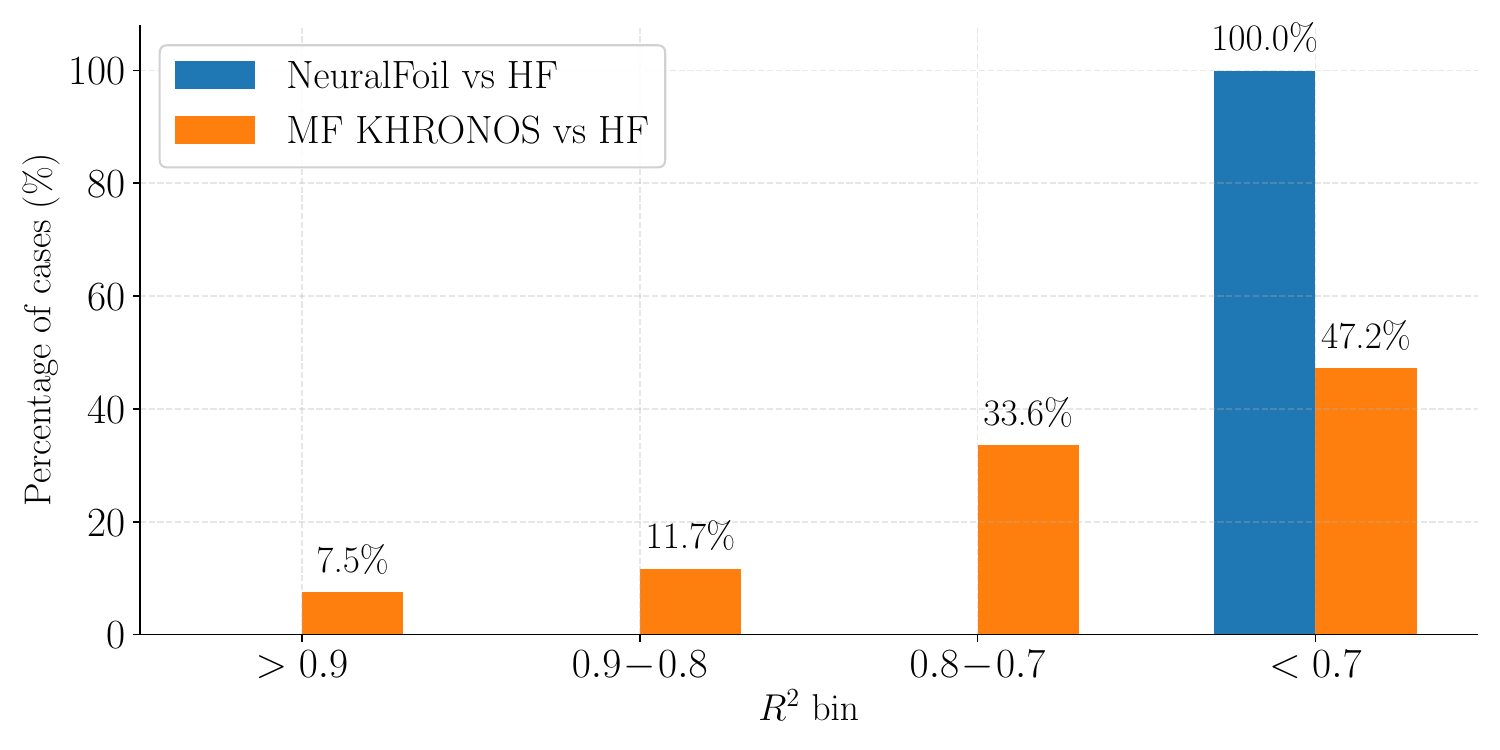}
  \caption{Distribution of coefficient of determination $R^{2}$ over the low accuracy dataset, binned into four ranges and compared between the \textit{NeuralFoil} low–fidelity model and the multi–fidelity KHRONOS surrogate, both evaluated against the high–fidelity (HF) data. The multi-fidelity gains can be observed here. \textit{NeuralFoil} predictions (blue bars) are concentrated in the $<0.7$ bin, whereas MF KHRONOS (orange bars) with the multi-fidelity gains shifts a substantial fraction of cases toward higher $R^{2}$}
  \label{fig:r2_bins_grouped}
\end{figure}

The overall impact of this behavior on the test set is summarized in Fig.~\ref{fig:r2_bins_grouped}. When $R^{2}$ values are binned over all cases, the \textit{NeuralFoil} LF model is found to reside exclusively in $<0.7$ bin, indicating that a majority of configurations with strong suction heads are poorly captured. In contrast, the MF KHRONOS model shifts a significant fraction of cases into higher–accuracy bins: approximately $52.8\%$ of the cases attain $R^{2} > 0.7$, and the proportion of very low–$R^{2}$ predictions ($<0.7$) is reduced compared with \textit{NeuralFoil}. These results indicate that KHRONOS remains accurate both for high–$R^{2}$ and low–$R^{2}$ LF regimes and that, in the latter, the learned correction is particularly effective at recovering the missing suction peaks associated with varied geometries and a wide range of $U$ and AoA.

\section{Conclusions}
\label{sec:conclusion}


This study employed KHRONOS, a kernel-based neural surrogate for multi-fidelity prediction, excelling in resource-constrained environments. Under strict training wall-clock budgets of around 15 seconds, KHRONOS achieves a prediction error (1-$R^2$) near 0.1, while baseline models still show noticeably larger errors. Across Cases 1--3 with increasing HF mixes, KHRONOS maintains comparable test accuracy while using 94--97\% and 96--98\% fewer parameters than MLP/PINN and GNN, respectively. Inference is faster by roughly 44--63\% relative to MLP and 52--62\% relative to GNN, and training time per fold drops by about 40--73\% versus MLP and 53--70\% versus GNN, depending on the fidelity mix. Beyond efficiency gains, KHRONOS also demonstrates strong corrective capability on the low-accuracy subset, where \textit{NeuralFoil}'s predictions fall entirely in the $R^2 < 0.7$ bin against HF, multi-fidelity KHRONOS shifts 52.8\% of cases to $R^2 > 0.7$. This suggests the learned $\Delta$ correction is most valuable precisely where the LF physics surrogate breaks down. However, the evaluation is limited to a single HF source (AirfRANS), a single LF generator (\textit{NeuralFoil}), and three HF mixing ratios. Future work will expand to additional HF datasets and LF surrogates, explore wider and adaptive HF/LF allocation policies, and test generalization to out-of-distribution airfoils and operating conditions.

\section*{Appendix}
\subsection{Hyperparameter Grid Search}
\label{sec:Appendix_grid_search}
A model-specific grid search is conducted to select hyperparameters for the surrogate and baseline models. For each model, the search space is defined as a discrete set of candidate values, and candidate configurations are generated through the Cartesian product of these sets. Table \ref{tab:grid_search_spaces} describes the search space.

\begin{table}[h!]
\centering
\small
\caption{Hyperparameter search spaces used for KHRONOS, GNN, MLP, and PINN}
\label{tab:grid_search_spaces}

\newcolumntype{Y}{>{\raggedright\arraybackslash}X}

\begin{tabularx}{\linewidth}{p{0.12\linewidth} p{0.36\linewidth} Y}
\hline
\textbf{Model} & \textbf{Hyperparameter} & \textbf{Values tested} \\
\hline

\multirow{10}{*}{KHRONOS}
& LF elements, $k_{\mathrm{elem}}$ & $\{3, 5, 8, 12\}$ \\
& LF rank, $k_{\mathrm{rank}}$ & $\{4, 5, 10, 15\}$ \\
& LF peak learning rate, $\eta_{\mathrm{LF}}$ & $\{0.001, 0.003, 0.005, 0.008\}$ \\
& LF epochs, $N_{\mathrm{LF}}$ & $\{1000, 1500, 2500, 4500\}$ \\
& $\Delta$ elements, $k_{\mathrm{elem}}$ & $\{3,4, 5, 8, 12, 15\}$ \\
& $\Delta$ rank, $k_{\mathrm{rank}}$ & $\{3, 4, 5, 6, 10, 15\}$ \\
& $\Delta$ peak learning rate, $\eta_{\Delta}$ & $\{0.001, 0.003, 0.005, 0.008\}$ \\
& $\Delta$ epochs, $N_{\Delta}$ & $\{1000, 1500, 3000, 6000\}$ \\
\hline

\multirow{8}{*}{GNN}
& Hidden dimension & $\{64, 96, 128, 256\}$ \\
& Number of message-passing layers & $\{2, 3, 4\}$ \\
& Graph connectivity ($k$-NN) & $\{2, 4, 8, 16\}$ \\
& LF learning rate, $\eta_{\mathrm{LF}}$ & $\{10^{-3}, 3\times10^{-3}, 5\times10^{-3}\}$ \\
& $\Delta$ learning rate, $\eta_{\Delta}$ & $\{5\times10^{-4}, 10^{-3}, 3\times10^{-3}\}$ \\
& Batch size & $\{512, 1024, 2048\}$ \\
& Activation & \{tanh, gelu, relu\} \\
& Dropout & $\{0.0, 0.1, 0.2\}$ \\
\hline

\multirow{11}{*}{MLP}
& LF architecture &
\begin{tabular}[t]{@{}l@{}}
\{(128,128,128), (256,256,256), (256,256,256,256),\\
(512,512,512), (256,256,256,256,256)\}
\end{tabular} \\

& $\Delta$ architecture & $\{(128,128,128), (128,128,128,128), (256,256,256,256)\}$ \\

& LF learning rate, $\eta_{\mathrm{LF}}$ & $\{10^{-3},  3\times10^{-3},  5\times10^{-3}, 10^{-2}\}$ \\
& $\Delta$ learning rate, $\eta_{\Delta}$ & $\{5\times10^{-4}, 10^{-3}, 5\times10^{-3}\}$ \\
& LF batch size & $\{1024, 2048, 4096\}$ \\
& $\Delta$ batch size & $\{256, 512, 1024\}$ \\
& Activation & \{tanh, relu, gelu\} \\
& LF epochs, $N_{\mathrm{LF}}$ & $\{1000, 2000, 3000, 3500\}$ \\
& $\Delta$ epochs, $N_{\Delta}$ & $\{1000, 1500, 2000, 4000\}$ \\
& Learning-rate decay rate & $\{0.9, 0.95, 0.99\}$ \\
& Learning-rate decay steps & $\{250, 500, 1000\}$ \\
\hline

\multirow{17}{*}{PINN}
& LF architecture &
\begin{tabular}[t]{@{}l@{}}
\{(128,128,128), (256,256,256), (256,256,256,256),\\
(512,512,512), (256,256,256,256,256)\}
\end{tabular} \\

& $\Delta$ architecture & $\{(128,128,128), (128,128,128,128), (256,256,256,256)\}$ \\
& LF learning rate, $\eta_{\mathrm{LF}}$ & $\{10^{-3}, 3\times10^{-3}, 5\times10^{-3}, 10^{-2}\}$ \\
& $\Delta$ learning rate, $\eta_{\Delta}$ & $\{5\times10^{-4}, 10^{-3}, 5\times10^{-3}\}$ \\
& LF batch size & $\{1024, 2048, 4096\}$ \\
& $\Delta$ batch size & $\{256, 512, 1024\}$ \\
& Activation (data-driven sweep) & \{tanh, relu, gelu\} \\
& LF epochs, $N_{\mathrm{LF}}$ & $\{1000, 2000, 3000, 3500\}$ \\
& $\Delta$ epochs, $N_{\Delta}$ & $\{1000, 1500, 2000, 4000\}$ \\
& Learning-rate decay rate & $\{0.9, 0.95, 0.99\}$ \\
& Learning-rate decay steps & $\{250, 500, 1000\}$ \\
& PDE loss weight, $\lambda_{\mathrm{PDE}}$ & $\{10^{-3}, 10^{-2}, 10^{-1}\}$ \\
& Far-field loss weight, $\lambda_{\mathrm{far}}$ & $\{10^{-3}, 10^{-2}, 10^{-1}\}$ \\

\end{tabularx}
\end{table}

\subsection{K-Fold Cross-Validation}

\label{sec:K-fold}

\begin{figure}[h!]
\centering
  \includegraphics[width=0.8\linewidth]{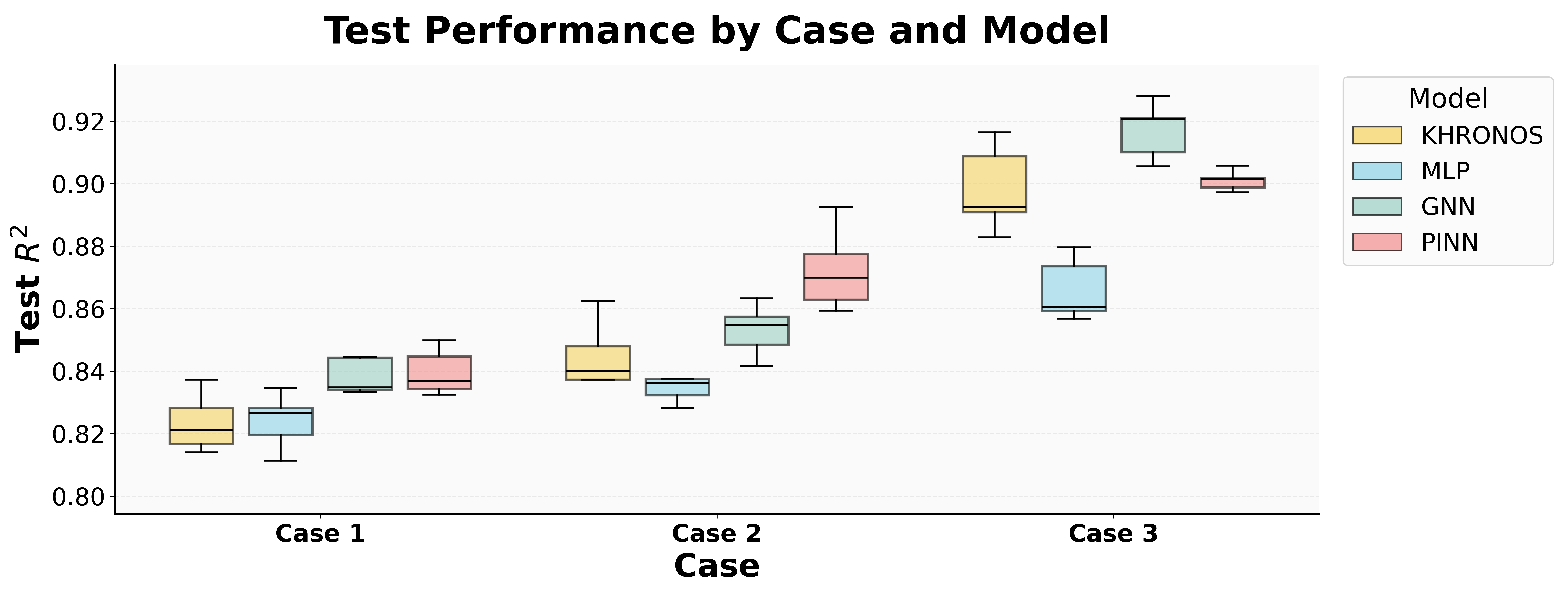}
  \caption{K-Fold Cross validation $R^2$ distribution}
  \label{fig:k-fold_R2}
\end{figure}

K-fold cross-validation is used to estimate generalization performance by repeatedly training and testing the model on different partitions of the same dataset. The full set of cases is first partitioned into 5 approximately equal-sized folds. In each iteration, one fold is held out as the test subset, and the remaining 4 folds are used for training. This procedure is repeated $K$ times so that each case appears in the test set exactly once, while being used for training in the other $K-1$ iterations. The performance metrics are then computed on the held-out fold for each iteration and aggregated across folds, commonly by averaging, to obtain a robust estimate of model performance that is less sensitive to a particular train-test split. In this implementation, $K=5$ folds are used with shuffling enabled and a fixed random seed to ensure reproducible splits. The statistical distribution for the Test $R^2$ values is shown in \ref{fig:k-fold_R2}. In the main text, only the mean values are reported.

\subsection{LF dataset summary of accuracy in terms of $R^2$}

\label{sec:neuralfoil-r2-bin}

\begin{figure}[h!]
\centering
  \includegraphics[width=0.8\linewidth]{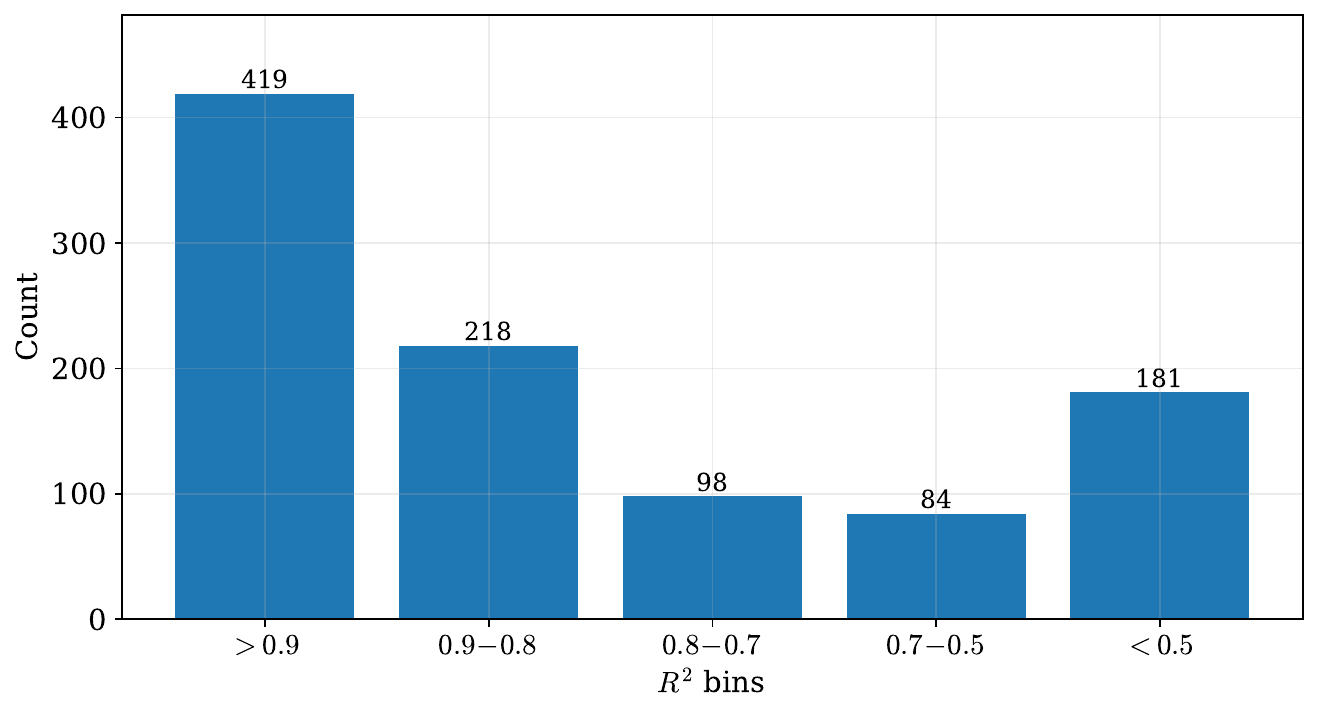}
  \caption{Distribution of Dataset across $R^2$ bins. Bars show the total count of cases in each $R^2$ range.}
  \label{fig:R2-bins}
\end{figure}

\textit{NeuralFoil's} predictive accuracy for $C_p$ from AirfRANS data varies with the geometric complexity and flow conditions. It is observed that 73.5\% of the cases are able to predict the AirfRANS $C_p$ with an accuracy of $R^2$ > 0.7. This limit is chosen to demarcate high-agreement. The remaining 26.5\% , the low-agreement cases, are used to demonstrate the performance gains of multi-fidelity surrogates compared with single-fidelity models. The number of cases in different $R^2$ bins is described in Figure \ref{fig:R2-bins}.


\newpage
\section*{Acknowledgments}
S.Saha gratefully acknowledges the start-up fund provided by the Kevin T. Crofton Department of Aerospace and Ocean Engineering, Virginia Tech, for supporting this work. The authors acknowledge helpful discussions with Mr. Kiavash Kamali (Politecnico di Milano, Italy) on geometry parameterization. 

\bibliography{references}

\end{document}